\documentclass{article}
\PassOptionsToPackage{numbers,sort&compress}{natbib}

\usepackage[final]{neurips_2024} 

\usepackage{float}
\usepackage[bottom]{footmisc}

\usepackage[utf8]{inputenc}
\usepackage[T1]{fontenc}
\usepackage[hidelinks]{hyperref}
\usepackage{url}
\usepackage{booktabs}
\usepackage{amsfonts}
\usepackage{nicefrac}
\usepackage{microtype}
\usepackage{graphicx}
\usepackage[dvipsnames]{xcolor}
\usepackage{amsmath}
\usepackage{amsthm}
\usepackage{amssymb}
\usepackage{mathrsfs}
\usepackage{marginnote}

\usepackage[capitalize,noabbrev]{cleveref}

\usepackage{tikz}
\usepackage{mathdots}
\usepackage{yhmath}
\usepackage{cancel}
\usepackage{color}
\usepackage{siunitx}
\usepackage{array}
\usepackage{multirow}
\usepackage{textcomp, gensymb}
\usepackage{gensymb}
\usepackage{extarrows}
\usetikzlibrary{fadings}
\usetikzlibrary{patterns}
\usetikzlibrary{shadows.blur}
\usetikzlibrary{shapes}
\usepackage{wrapfig}
\usepackage{bm}
\usepackage{xspace}
\usepackage{capt-of}

\usepackage{todonotes}

\let\originalleft\left
\let\originalright\right
\renewcommand{\left}{\mathopen{}\mathclose\bgroup\originalleft}
\renewcommand{\right}{\aftergroup\egroup\originalright}

\newcommand{\tr}{\textsf{T}}
\DeclareMathAlphabet{\mat}{OT1}{cmss}{bx}{n}

\newtheorem*{proposition*}{Proposition}

\newcommand{\dt}[0]{Deep Thinking}
\newcommand{\DT}[0]{DT}
\newcommand{\dtr}[0]{Deep Thinking with Recall}
\newcommand{\DTR}[0]{DT-R}
\newcommand{\dtl}[0]{Deep Thinking with Lipschitz Constraints}
\newcommand{\DTL}[0]{DT-L}

\newcommand{\networkF}{\ensuremath{\mathcal{F}}\xspace}
\newcommand{\networkG}{\ensuremath{\mathcal{G}}\xspace}
\newcommand{\networkH}{\ensuremath{\mathcal{H}}\xspace}

\ifluatex
\newcommand{\bm}{\symbfup}
\else
  \usepackage{bm}
  \newcommand{\symbf}[1]{\bm{#1}}
  
\fi

\def\vphi{{\symbf{\phi}}}

\def\vc{{\symbf{c}}}

\def\vv{{\symbf{v}}}
\def\vw{{\symbf{w}}}
\def\vx{{\symbf{x}}}
\def\vy{{\symbf{y}}}

\newcommand{\p}{p}
\newcommand{\q}{q}
\newcommand{\probsym}{\q}

\newcommand{\true}[1]{
\global\let\probsym\p
#1
\global\let\probsym\q
}

\newcommand{\norm}[1]{\left\| #1 \right\|}

\title{Rethinking Deep Thinking: Stable Learning of Algorithms using Lipschitz
Constraints}

\author{
    Jay Bear \quad\quad\quad Adam Pr\"{u}gel-Bennett \quad\quad\quad Jonathon Hare\\
    The University of Southampton, Southampton, UK\\
    \texttt{\{jdh1g19,abp1,jsh2\}@soton.ac.uk}
}

\begin{document}

\maketitle

\begin{abstract}
    Iterative algorithms solve problems by taking steps until a solution is reached. Models in the form of \dt{} (\DT{}) networks have been demonstrated to learn iterative algorithms in a way that can scale to different sized problems at inference time  using recurrent computation and convolutions. However, they are often unstable during training, and have no guarantees of convergence/termination at the solution. This paper addresses the problem of instability by analyzing the growth in intermediate representations, allowing us to build models (referred to as \dtl{} (\DTL{})) with many fewer parameters and providing more reliable solutions. Additionally our \DTL{} formulation provides guarantees of convergence of the learned iterative procedure to a unique solution at inference time. We demonstrate \DTL{} is capable of robustly learning algorithms which extrapolate to harder problems than in the training set. We benchmark on
    the traveling salesperson problem to evaluate the capabilities of the
    modified system in an NP-hard problem where \DT{} fails to learn.%
\end{abstract}

\section{Introduction}

Iteration is a key ingredient in a vast number of important algorithms. Incremental progress
towards a solution is demonstrated in many of these. Well-known examples include insertion
sort, gradient descent, and the simplex method. This paper explores models that learn iterative algorithms. We ask questions including: Can deep learning models be
used to learn the complex steps of algorithms similar to these? Is there a way
to guarantee a solution or approximation is reached? Can we learn algorithms that extrapolate to larger or harder instances than we train on? In addressing these questions we propose a model called \dtl{} (\DTL{}).
\medbreak

\begin{center}
    \centering
    \resizebox{0.8\textwidth}{!}{

\tikzset{every picture/.style={line width=0.75pt}} 

\begin{tikzpicture}[x=0.75pt,y=0.75pt,yscale=-1,xscale=1]

\draw   (144,47) -- (164,47) -- (164,67) -- (144,67) -- cycle ;
\draw    (164,57) -- (191,57) ;
\draw [shift={(194,57)}, rotate = 180] [fill={rgb, 255:red, 0; green, 0; blue, 0 }  ][line width=0.08]  [draw opacity=0] (7.14,-3.43) -- (0,0) -- (7.14,3.43) -- (4.74,0) -- cycle    ;
\draw   (194,47) -- (214,47) -- (214,67) -- (194,67) -- cycle ;
\draw   (20.5,47) -- (38,47) -- (30.5,67) -- (13,67) -- cycle ;
\draw    (119,52) -- (141,52) ;
\draw [shift={(144,52)}, rotate = 180] [fill={rgb, 255:red, 0; green, 0; blue, 0 }  ][line width=0.08]  [draw opacity=0] (7.14,-3.43) -- (0,0) -- (7.14,3.43) -- (4.74,0) -- cycle    ;
\draw  [draw opacity=0][dash pattern={on 4.5pt off 4.5pt}] (164.34,31.88) .. controls (166.94,32.05) and (169,34.22) .. (169,36.87) .. controls (169,36.91) and (169,36.96) .. (169,37) -- (164,36.87) -- cycle ; \draw  [dash pattern={on 4.5pt off 4.5pt}] (164.34,31.88) .. controls (166.94,32.05) and (169,34.22) .. (169,36.87) .. controls (169,36.91) and (169,36.96) .. (169,37) ;  
\draw  [dash pattern={on 4.5pt off 4.5pt}]  (169,56.87) -- (169,37) ;
\draw  [dash pattern={on 4.5pt off 4.5pt}]  (164.34,31.88) -- (124,32) ;
\draw  [draw opacity=0] (119.01,36.66) .. controls (119.19,34.06) and (121.35,32) .. (124,32) .. controls (124.04,32) and (124.09,32) .. (124.13,32) -- (124,37) -- cycle ; \draw   (119.01,36.66) .. controls (119.19,34.06) and (121.35,32) .. (124,32) .. controls (124.04,32) and (124.09,32) .. (124.13,32) ;  
\draw  [dash pattern={on 4.5pt off 4.5pt}]  (119,42.13) -- (119,49) ;
\draw [shift={(119,52)}, rotate = 269.99] [fill={rgb, 255:red, 0; green, 0; blue, 0 }  ][line width=0.08]  [draw opacity=0] (7.14,-3.43) -- (0,0) -- (7.14,3.43) -- (4.74,0) -- cycle    ;
\draw    (214,57) -- (230,57) ;
\draw [shift={(233,57)}, rotate = 180] [fill={rgb, 255:red, 0; green, 0; blue, 0 }  ][line width=0.08]  [draw opacity=0] (7.14,-3.43) -- (0,0) -- (7.14,3.43) -- (4.74,0) -- cycle    ;
\draw   (236.5,47) -- (254,47) -- (246.5,67) -- (229,67) -- cycle ;
\draw   (64,27) -- (84,27) -- (84,47) -- (64,47) -- cycle ;
\draw    (84,37) .. controls (114.5,37.33) and (84.5,52) .. (119,52) ;
\draw    (34,57) .. controls (58.61,57) and (36.19,38.38) .. (61.1,37.07) ;
\draw [shift={(64,37)}, rotate = 180] [fill={rgb, 255:red, 0; green, 0; blue, 0 }  ][line width=0.08]  [draw opacity=0] (7.14,-3.43) -- (0,0) -- (7.14,3.43) -- (4.74,0) -- cycle    ;
\draw  [dash pattern={on 0.84pt off 2.51pt}]  (34,57) .. controls (59.37,57.33) and (9.67,62.28) .. (141.99,62) ;
\draw [shift={(144,62)}, rotate = 179.86] [fill={rgb, 255:red, 0; green, 0; blue, 0 }  ][line width=0.08]  [draw opacity=0] (7.14,-3.43) -- (0,0) -- (7.14,3.43) -- (4.74,0) -- cycle    ;

\draw (237,53.4) node [anchor=north west][inner sep=0.75pt]  [font=\scriptsize]  {$\vy$};
\draw (85,23.4) node [anchor=north west][inner sep=0.75pt]  [font=\scriptsize]  {$\vphi ^{(0)}$};
\draw (198,52) node [anchor=north west][inner sep=0.75pt]  [font=\scriptsize]  {$\networkH$};
\draw (68.2,31.9) node [anchor=north west][inner sep=0.75pt]  [font=\scriptsize]  {$\networkF$};
\draw (149.5,52) node [anchor=north west][inner sep=0.75pt]  [font=\scriptsize]  {$\networkG$};
\draw (156,18.4) node [anchor=north west][inner sep=0.75pt]  [font=\scriptsize]  {$\vphi^{\left(m+1\right)}$};
\draw (20.5,53.4) node [anchor=north west][inner sep=0.75pt]  [font=\scriptsize]  {$\vx$};
\draw (168.5,42) node [anchor=north west][inner sep=0.75pt]  [font=\scriptsize]  {$\vphi^{\left(\!M\!\right)}$};
\draw (125,37) node [anchor=north west][inner sep=0.75pt]  [font=\scriptsize]  {$\vphi^{\left(m\right)}$};

\end{tikzpicture}}
    
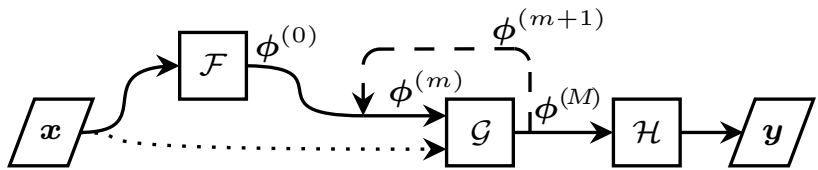
\captionof{figure}{Recurrent-based model architectures for learning algorithms with input $\vx$ and output $\vy$. $\networkF$, $\networkG$ and $\networkH$ are convolutional networks that work on any size input. A scratchpad $\vphi$ serves as the working memory during computation. As described in \cref{section:related} the original \DT{} model didn't include \textit{recall}, denoted by the dotted line. The improved \DTR{} and our \DTL{} model include this connection.}
    \label{fig:model-high-level}
\end{center}

Some work towards answering the above questions has occurred in recent years. A key approach has been to utilize various forms of a learned recurrent function to implement a `step' in an iterative algorithm that computes a solution to a problem. Typically these recurrent functions are combined with learned functions that pre-process the initial input, and post-process the output of the last iteration into the final solution form. Examples of such approaches are the \dt{} (\DT{}) networks~\citep{schwarzschild2021algorithm} and \dtr{} (\DTR{}) networks~\citep{bansal2022endtoend} that we build upon in this work (see \cref{fig:model-high-level}).

There are two key features to these models: firstly the use of an iteratively updated \textit{scratchpad}, \(\vphi\) serving as a working memory; and secondly, the construction of the model from convolutional layers to enable it to scale to arbitrary sized problems. The latter is important because in principle it allows training on smaller problems, and then extrapolation at inference time to larger problems~\citep{schwarzschild2021algorithm, bansal2022endtoend}.

Many challenges still remain with the existing \DT{} and \DTR{} approaches however, and we aim to address these in this work. Firstly, \DT{}-style networks are quite difficult to train and can be very unstable both during training and inference. The \DT{}-style networks previously demonstrated in the literature are massively overparameterized --- we show in \cref{section:dt-analysis} that model width is inversely related to training stability in existing models, and later demonstrate how this can be addressed (see \cref{section:refinement}). Secondly, existing approaches have no guarantees on the convergence of the learned algorithm; this is important because if a solution to a problem exists, one would hope that an algorithm to solve it would terminate or reach a stable state once the solution has been reached. \cref{section:refinement} details a design approach to guarantee convergence within the limits of floating point precision by considering the network in terms of a discrete dynamical system.
Our contributions are:
\begin{itemize}
    \item We systematically analyze \DT{} networks and explore why they can be
          unstable;
    \item we introduce techniques grounded in theory into the \DT{} framework
          which improve the stability of learning and guarantee convergence at run time. This allows the
          use of much smaller models to tackle the same problems, and improves
          extrapolation performance;
    \item we propose our
          \dtl{} (\DTL{}) model and perform a comprehensive evaluation and ablation study; and,
    \item we show that the approach can be applied to learn to find low-cost
          tours for a range of TSPs, including non-Euclidean and
          asymmetric instances.
\end{itemize}

\section{Related work}\label{section:related}

\dt{} (\DT{}) networks~\citep{schwarzschild2021algorithm} were designed to learn algorithms by using
recurrence to induce iterative behavior in such a way they could be trained on smaller and simpler examples before extrapolating to larger tests. The networks were used to solve `Easy-to-Hard' problems (consisting of prefix sums, mazes, and chess puzzles \citep{schwarzschild2021dataset}) and they could be trained with few iterations on easy problems in such a way that they can solve harder problems by increasing the number of iterations.

\DT{} networks were further improved by \citet{bansal2022endtoend} by means of
adding `recall' (which we refer to as \DTR{} networks) --- a mechanism for the recurrent component to have continuous
access to the original input. Recall successfully allowed \DTR{} networks to solve
much larger tests than \DT{} networks without recall, while also mitigating
\textit{overthinking}, where if after too many iterations at inference time the predicted solution becomes progressively worse.
In addition to recall, \citeauthor{bansal2022endtoend} introduced
\textit{incremental progress training} (IPT), a training method which disables gradient tracking for a random number of initial iterations. This prevents the model from learning behaviors based strictly on the number of iterations and instead promotes incremental modification to the internal states.

There exist alternative approaches to producing machines which can learn
algorithms. One set of models are Neural Turing Machines (NTMs), which use
attention in a recurrent system for reading to, and writing from, some external
memory \citep{graves2014turing}. NTMs showed success in learning cell-based
copying and repeating tasks which could be applied to unseen inputs.
Differentiable Neural Computers (DNCs) operate similarly to NTMs, differing
primarily in their method of external memory access \citep{graves2016neural}. In the NTM, like most RNNs, the amount of compute is directly tied to the input sequence length. \citet{DBLP:journals/corr/Graves16} proposed a method to adaptively select the compute budget in RNNs. The models studied here are different in the sense that the objective is to extrapolate beyond the training data to harder problem instances as the compute budget is increased, and the input and output is not a sequence. 

A second class of models exist where a specific base iterative algorithm is defined, and the parameters of the algorithm, or the problem, are learned. Examples include the reverse diffusion process learned in diffusion models~\citep{NEURIPS2020_4c5bcfec}, and models which involve optimization using gradient descent iterations within the forward pass to generate an output with certain properties that are difficult to produce directly with a neural network, such as permutation equivariance~\citep{NEURIPS2019_6e79ed05}. Our work does not constrain the class of algorithm used directly, other than enforcing that we learn an iterative one.

\section{Analysis of \dt{} Networks}\label{section:dt-analysis}

The main architecture used in pre-existing \dt{} networks~\citep{schwarzschild2021algorithm, bansal2022endtoend}, and our own network, is shown in \cref{fig:model-high-level}.  It takes an input problem instance \(\vx\) and generates a solution \(\vy\).  The model consists of three convolutional networks \(\networkF\), \(\networkG\) and \(\networkH\). The function \(\networkF\) is responsible for initially pre-processing the input \(\vx\) into the initial state \(\vphi^{\left(0\right)}\); function \networkG is the recurrent function that takes the current state, \(\vphi^{\left(m\right)}\) (plus the original input \(\vx\) in the case of \DTR{} and our models), to produce the next state, \(\vphi^{\left(m+1\right)}\). The final state produced by \networkG, after \(M\) iterations, is denoted \(\vphi^{\left(M\right)}\). The function \networkH takes \(\vphi^{\left(M\right)}\) and produces the predicted output \(\vy\).  Formally
\begin{align}
    \vphi^{\left(0\right)} &= \networkF\left(\vx\right) \;,\\
    \vphi^{\left(m+1\right)} &= \networkG\left(\vphi^{\left(m\right)}, \vx\right) \quad \forall m \in \left\{0,\ldots, M-1\right\}\;,
    \label{eq:iteration}\\
    \vy &= \networkH\left(\vphi^{\left(M\right)}\right)\;.
\end{align}
Architecturally these networks are recurrent neural networks. Unlike more commonly used recurrent networks such as LSTMs they are not used to tackle problems with sequential data, but rather the recurrent part is used to find a solution through an iterative process.  The \textit{algorithm} that the recurrent network uses, as well as the pre- and post- processing functions are learned through supervised training on example input-solution pairs for a particular problem.

As the networks \networkF, \networkG and \networkH are convolutional neural networks they can work with an arbitrary size input.  The solution $\bm{y}$ is typically the same size as the input $\bm{x}$. The feature of \DT{} and \DTR{} that excited interest was that, not only could they solve unseen problem instances of the same size as they were trained on (which we call \textit{interpolation}), but when trained on small problem instances (with relatively small $M$), they were able to find low cost solutions on much larger problem instances (which we refer to as \textit{extrapolation}) by increasing $M$ at inference time.
Both \DT{} and \DTR{} suffer from poor stability both in training but particularly at inference time when extrapolating to larger problems.  This often lead to overflow errors.  This was particularly seen if the number of channels in network \networkG is reduced, then \DT{} and \DTR{} are hard to train with many runs failing to find a solution.  We show the reason for this is that there is no mechanism to control the change in size of the scratchpad representation, $\|\vphi^{(m)}\|/\|\vphi^{(m-1)}\|$, leading to this either overflowing or vanishing during learning.

\subsection{Training Stability}

\begin{figure}[b]
    \centering
    \includegraphics[scale=0.3]{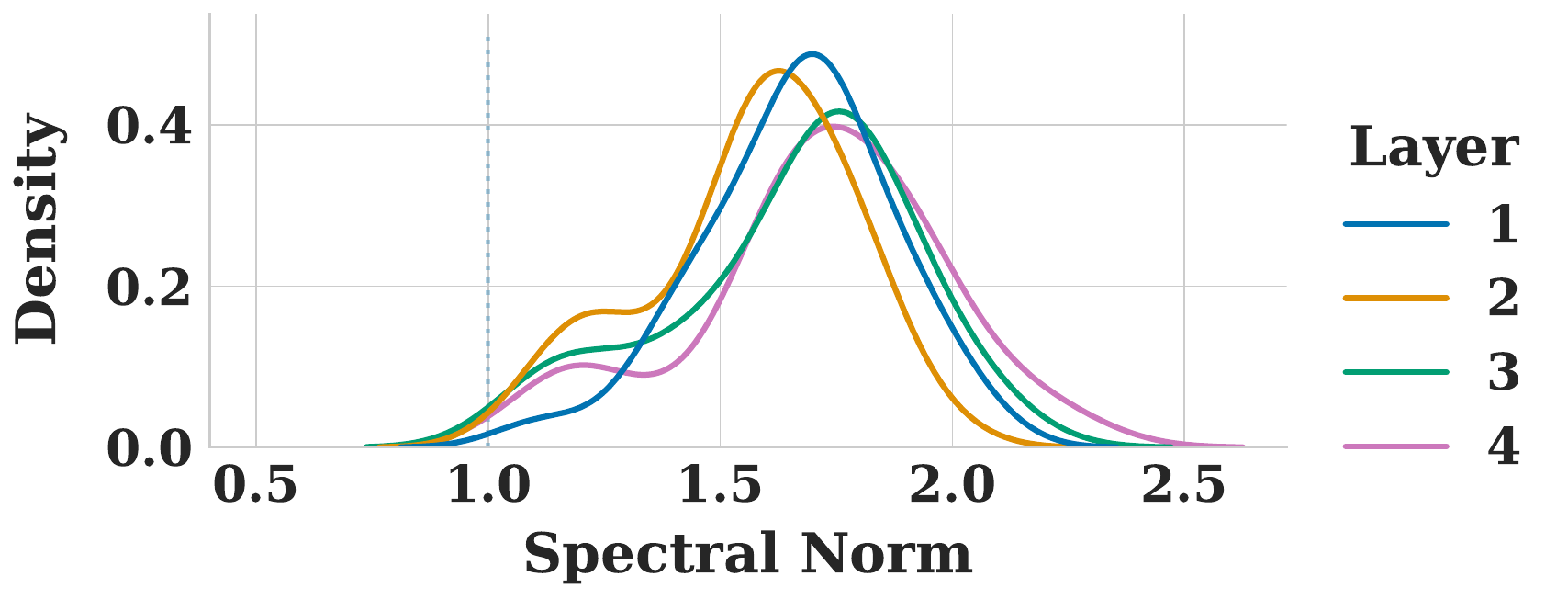}
    \captionof{figure}{Distribution of spectral norms of reshaped weight matrices for the different convolutional layers in the recurrent part of
    \DTR{}. 30 prefix-sum-solving models with width \(w=32\) were sampled.}
    \label{fig:dtanalysis:spectral}
\end{figure}

\begin{figure}[t!]
    \centering
    \begin{tabular}{ccc}
        \includegraphics[scale=0.25]{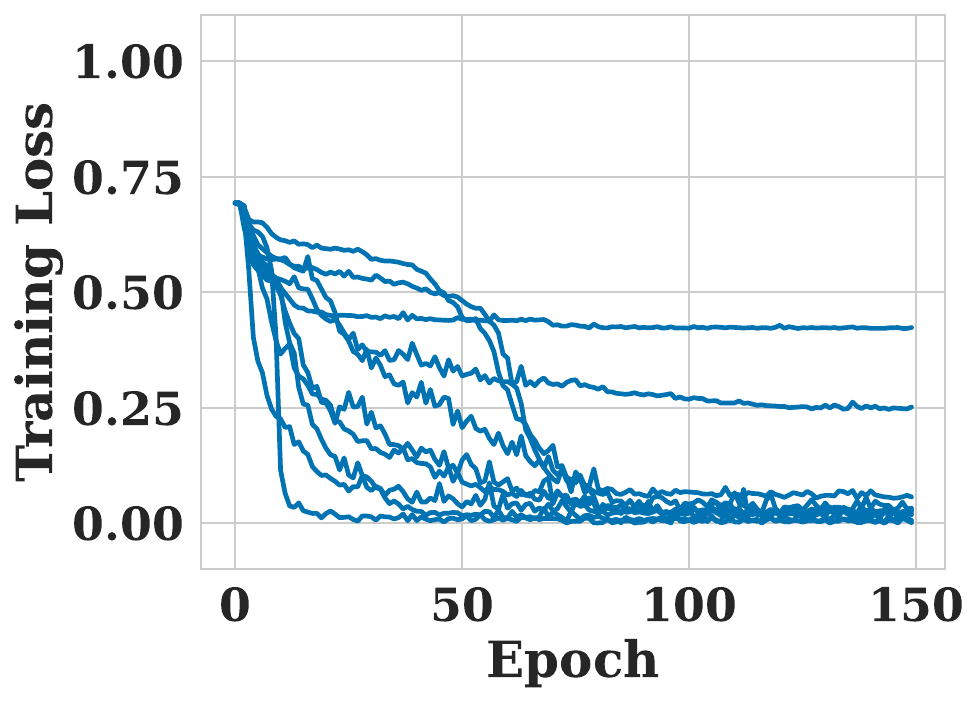}&
        \includegraphics[scale=0.25]{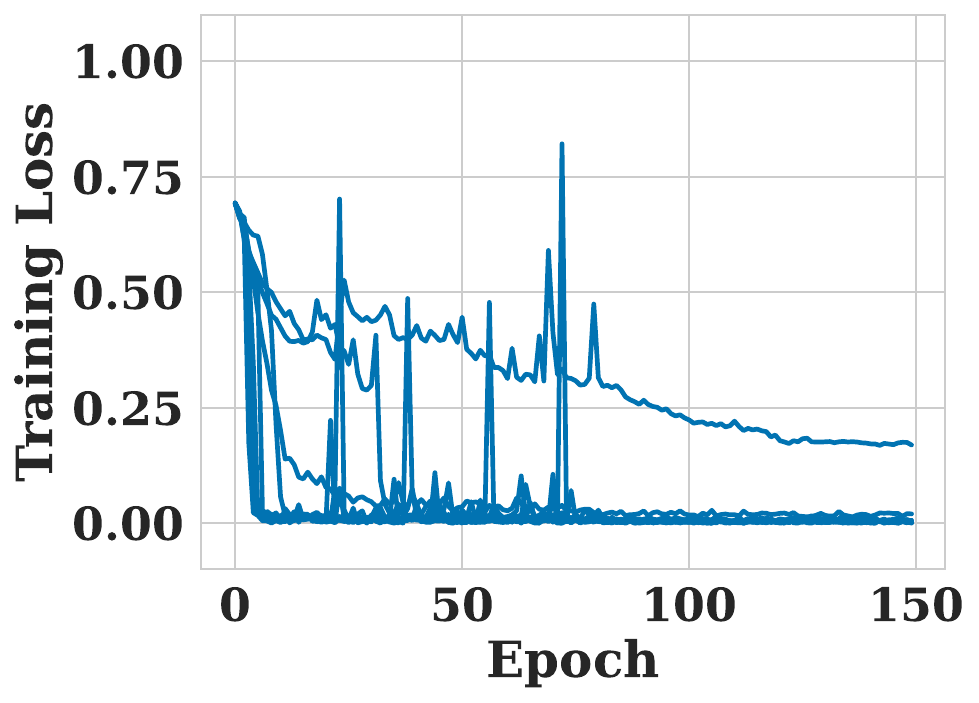}&
        \includegraphics[scale=0.25]{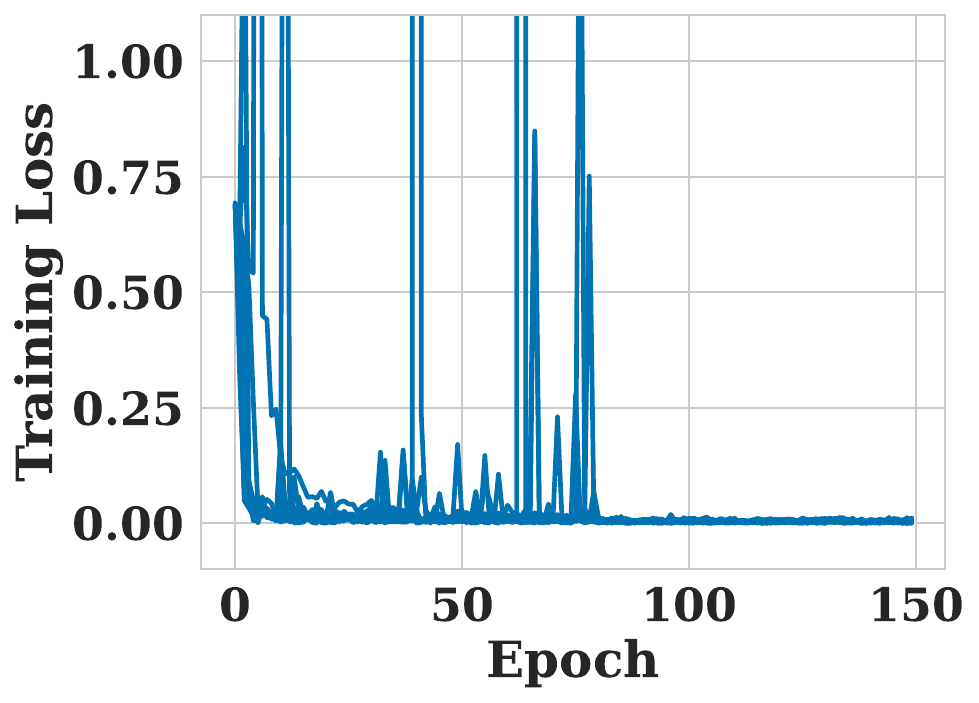}\\
        \(w=16\)&\(w=64\)&\(w=256\)
    \end{tabular}
    \captionof{figure}{Mean training (cross-entropy) loss at each epoch for
                       prefix-sums-solving models of varying width \(w\). For small $w$ training is stable, but not all models converge; larger $w$ has a higher chance of models reaching a small loss, but the training process has very large spikes in the loss which causes some models to explode. Each
                       curve is measured from a different random
                       initialization of the model throughout training, for
                       10 models of each width}
    \label{fig:dtanalysis:stability}
\end{figure}

In this section we focus on \DTR{} as this is more stable than the original \DT{} network. We use the \textit{spectral norm} of the reshaped weights of a convolution\footnote{Given a convolutional layer with weights shaped $(C_{out}, C_{in}, n)$, where $n$ is the number of elements (e.g. the kernel height times width in a 2D convolution) we can flatten the weights into a matrix of shape $C_{out} \times C_{in} \cdot n$; the spectral norm is the largest singular value of this matrix.} to capture the expansion or shrinkage in magnitude of the output relative to the input of a sub-network, with a value of 1 meaning that the  magnitude is constant. In \cref{fig:dtanalysis:spectral}, we use violin plots to show the distribution of spectral norms for the four convolutional layers that arise after training \DTR{} on the prefix-sum problem (see \cref{sec:results} for a description of the problem). We observe that the spectral norms are typically greater than one, which can lead to the norm of $\vphi^{(m)}$ growing with each iteration.  In turn this can cause the model to overflow when applying the model to a larger input (the extrapolation scenario), where we increase the iteration number $M$ in order to solve the larger instance size. The \DT{} models as described and implemented by \citet{bansal2022endtoend} have training behavior which becomes increasingly unpredictable as the width \(w\) (number of channels in the scratchpad) is reduced. This can be seen in \cref{fig:dtanalysis:stability}, where the variation in training loss at the end of training over different runs is larger in models of smaller width. It is also worth noting that increasing the width can result in explosive behavior in loss, including not-a-number (NaN) results, which can be seen in \cref{fig:dtanalysis:stabilityext}.

\subsection{Extrapolation Performance}

The property of the \DT{} networks that excited our interest was their ability to solve large instances than they were trained (that is, their extrapolative ability). This generalization to out-of-distribution examples is a key challenge for many machine learning algorithms.  However, although \DTR{} can find models that extrapolate, very often the models that are trained extrapolate poorly.
We observe models trained on few recurrences may have explosive behavior which only
becomes a problem with extrapolative tasks requiring more recurrences.

Both \DT{} and \DTR{} network used no bias terms in their model \cite{schwarzschild2021algorithm, bansal2022endtoend}.  In experimenting with these networks, we found that adding bias made these models very unstable.  Unfortunately, having no biases exacerbates the `dead cell problem', where some of the convolutional filters would have a zero response to all training inputs.

\section{Refinement of the Architecture: \dtl{}}\label{section:refinement}

In the previous section we have identified a problem faced by \DTR{}, namely that there is instability in reaching a solution through the iteration in \cref{eq:iteration}.
To overcome this difficulty we use a well known property of contraction maps. If $c : \mathcal{V} \rightarrow \mathcal{V}$ is a mapping of objects, $\vv$ and $\vw$, in a normed vector space $\mathcal{V}$, that is contractive in the sense that
\begin{align}
    \| c(\vv) - c(\vw) \| < \| \vv - \vw \| \;,
\end{align}
implying $c(\cdot)$ is a Lipschitz function with Lipschitz constant $K<1$, then the iterations
\begin{align}
    \vv^{(m)} = c(\vv^{(m-1)})
\end{align}
will converge to a unique solution as a consequence of the Banach fixed-point theorem~\citep{banach,10.15352/bjma/1240321550}.  We can use this to refine \DTR{} by engineering the network $\networkG(\cdot, \vx)$ to be a contraction mapping. Noting that in \cref{eq:iteration} the input $\vx$ is held constant throughout the iteration, we construct $\networkG(\cdot, \vx)$ so that it has an \textit{approximate} Lipschitz constant $K$ less than 1.

Although any Lipschitz constant less than 1 would guarantee convergence, the nature of the problem solving mechanism we seek to learn intuitively means that we do not want fast convergence. This is because the network \networkG performs local convolutions with a limited receptive field. However, finding good quality solutions often requires a global knowledge. To accumulate such knowledge requires long distance information to accumulate in the scratchpad vector $\vphi^{(m)}$ over multiple iterations. To allow this to happen we choose the approximate Lipschitz constant associated with $\networkG$ to be less than, but close to 1.

\subsection{Constraining the Lipschitz Constant}

The primary tool we use to control the Lipschitz constant of $\networkG(\cdot, \vx)$ is spectral normalization~\citep{miyato2018spectral}. This is an extension of spectral norm regularization, introduced by \citet{yoshida2017spectral}, but it sets the spectral norm of an operator to a fixed value rather than penalizing the norm of an operator that differs from a predefined value. The method uses a power iteration to compute an approximation of the spectral norm of an operator, updated after each gradient step.  Computation of the spectral norm is a costly operation, but it only has to be done at the beginning of the iterative process.

As network \networkG consists of convolutions we divide each convolution kernel weight by the spectral norm plus a small constant $\varepsilon$ to ensure the spectral norm is less than one; there is one exception in that the convolution applied to the recall connection $\vx$ does not need normalizing --- please see \cref{app:architecture} for full details.  
In addition to using the spectral norm we need to make sure all the other transformations carried out by \networkG are at most 1-Lipschitz. However, to
avoid applying constraints to the constant recall connection, we replace \DTR{}'s method
of applying a single convolution on the concatenation
\(\left[\vphi^{\left(m\right)},\vx\right]\) with two separate convolutional
layers on \(\vphi^{\left(m\right)}\) and \(\vx\) respectively whose output is
combined by element-wise addition. See \cref{app:proofs} for a rigorous discussion on how the Lipschitz behaviour we require is ensured.

\paragraph{Constraining Activation Functions.}
To guarantee convergence under the constraints provided, any activation
function used in \networkG must be 1-Lipschitz.
This includes Rectified Linear Units (ReLUs), Exponential Linear Units (ELUs)
\citep{clevert2016elu}, and $\tanh$, but excludes Gaussian Error Linear Units
(GELUs) \citep{hendrycks2016gelu}, which have an absolute gradient greater
than 1 around \(x=\sqrt{2}\). Results in the body of this paper use the ELU activation for \DTL{} networks and the ReLU for \DTR{} as originally defined. Additional results using ReLU for \DTL{} and ELU for \DTR{} can be found in \cref{app:results:ablations}.

\paragraph{Constraining Residual Skip Connections.}
\DT{} and \DTR{} use element-wise addition for residual connections in
recurrent blocks. The sum of two functions
\(\operatorname{c}:X\to Y\) and \(\operatorname{d}:X\to Y\) applied to the same input, with Lipschitz
constants \(K_{\operatorname{c}}\) and \(K_{\operatorname{d}}\) respectively results in the upper bound,
\begin{equation}
    \left\|\left(\operatorname{c}\left(\vx_1\right)+\operatorname{d}\left(\vx_1
    \right)\right)-\left(\operatorname{c}\left(\vx_2\right)+\operatorname{d}
    \left(\vx_2\right)\right)\right\|\leq\left(K_{\operatorname{c}}+K_{
    \operatorname{d}}\right)\left\|\vx_1-\vx_2\right\|
\end{equation}
for all \(\vx_1,\vx_2\in X\). In the case of a standard residual connection, one
function is the identity (\(\operatorname{id}\left(\vx\right)=\vx\)) and the other is
the block of layers $\mathcal{B}$ contained in the span of the residual connection. Under our
constraints, the identity is 1-Lipschitz and the block of layers is
\(K_{\mathcal{B}}\)-Lipschitz, where \(K_{\mathcal{B}}\in\left[0,1
\right)\). This results in a Lipschitz upper bound for the output of
\(1+K_{\mathcal{B}}\).

As a result, residual connections as addition between the identity and a block of layers can increase the Lipschitz constant of the recurrent part even if the
layers themselves are 1-Lipschitz. A solution to this, which also allows more
expression in the model, is to make each residual connection a parametric
linear interpolation between the identity output and the block output
\begin{equation}
    \left(1-\gamma\right)\, \operatorname{id}(\vx)+\gamma\, \mathcal{B}\left(\vx\right), \quad
    \gamma\in\left[0,1\right] \;. \label{eqn:parametric_residual}
\end{equation}
\DTL{} applies this interpolation to each channel, $c$, individually; an unconstrained learnable parameter
\(\bar{\gamma_c}\) exists for each channel in each residual connection, then the residual parameters are set to
\(\gamma_c=\sigma\left(\bar{\gamma_c}\right)\) where \(\sigma\) is the logistic
function to ensure that $0\leq \gamma_c \leq 1$.

\subsection{Additional Modifications}

The above changes leads to a more stable network, allowing us to make additional modifications and still obtain convergent behaviour with improved performance. Without explicitly controlling the Lipschitz constant, these changes often lead to complete failure of the network to solve the problem.
In our final model design we apply three extra modifications that lead to consistent improvements:
\begin{enumerate}
    \item Batch normalization layers~\citep{sergey2015batchnorm} are added for the input and after each convolution,
          except for layers in the recurrent block and the final output layer. Empirically this improves performance.
    \item A bias term is added to the recall convolution (without the Lipschitz constraint) in \(\networkG\). The bias term is added to the recall convolution to mitigate the dead
cell problem. Since this is a constant addition (along with recall) it does not
disrupt the convergence constraints we have put in place (see \cref{app:proofs}). A bias term is also added to the
          final (output) layer and, when batch normalization is not used, to \(\networkF\).

    \item Exponential Linear Unit (ELU) activations
          \citep{clevert2016elu} are used instead of rectified linear unit (ReLU)
          activations. This choice is influenced by the desire to promote
activations staying close to zero \citep{clevert2016elu} throughout iteration,
as well as mitigating the effect of the dying ReLU problem 
\citep{trottier2016pelu} where `dead' activations increase with
depth \citep{lu2020dying}. Artificially deep models like \DT{} are therefore
likely to encounter this problem. Using ReLU can also result in a rapid rate
of convergence, preventing the model from learning complex algorithms for
harder tasks (see \cref{app:results:ablations}).
\end{enumerate}

Our reasoning for not applying batch normalization in the recurrent block
follows from~\citet{jastrzebski2018residual} where unsharing batch
normalization statistics was necessary to avoid activations exploding. In an
architecture where the maximum number of iterations is unknown it appears
infeasible to unshare batch normalization statistics for every iteration.

Using these modifications --- creating a model we call \dtl{} (\DTL) --- greatly improves the stability of training and using these networks. As we will see in \cref{sec:results}, the new \DTL{} network allow us to solve the same problems explored by \DT{} and \DTR{}, but using, in some cases, three orders of magnitude less parameters while achieving high performance and a more consistent success rate during training. The increased reliability offered by \DTL{} allows us to explore many other modifications enabling us to tackle more sophisticated problems. We illustrate this by attempting to find good solutions to one of the most notoriously difficult problems, namely, the traveling salesperson problem (TSP).
\section{Results on Easy-to-Hard Problems}
\label{sec:results}

In this section, we compare our model (\DTL) against \DTR{}. Note that \DTR{} is an improvement of \DT{} by the same authors and has already been shown to have better performance~\citep{bansal2022endtoend}.  We test on the three problem classes used by \citet{bansal2022endtoend} to evaluate \DTR{}, namely a \textbf{prefix sum problem}, a \textbf{maze problem} and a \textbf{chess problem}. 

We have trained and evaluated the models on a range of different Nvidia GPU accelerators from RTX2080Tis to A100s, as well as on M3-series Apple Silicon. Memory usage is insignificant compared to available GPU memory. Training time is of the order of 30 minutes for the $w=32$ model on the prefix-sum problem using a single RTX8000. More details can be found in \cref{app:runtimes}, and
code for the experiments can be found at
\url{https://github.com/Jay-Bear/rethinking-deep-thinking}.

The accuracy given in this section measures the proportion of instances where the network produces the exact correct solution. Note that even if a predicted solution differs from the target by one element (e.g. 1 bit in prefix sums, 1 pixel in mazes) it is considered a failure. 

\paragraph{Prefix Sums.} The prefix sum problem involves translating a string of ones and zero to a new string that counts sequences of ones (details are described by \citet{bansal2022endtoend}). The problem is simpler than the the maze and chess problem, and considerably faster to run.  As a consequence we choose this as the main problem to perform a comparative study.

\begin{figure}[ht!]
    \centering
    \begin{tabular}{ccc}
        \includegraphics[scale=0.325]{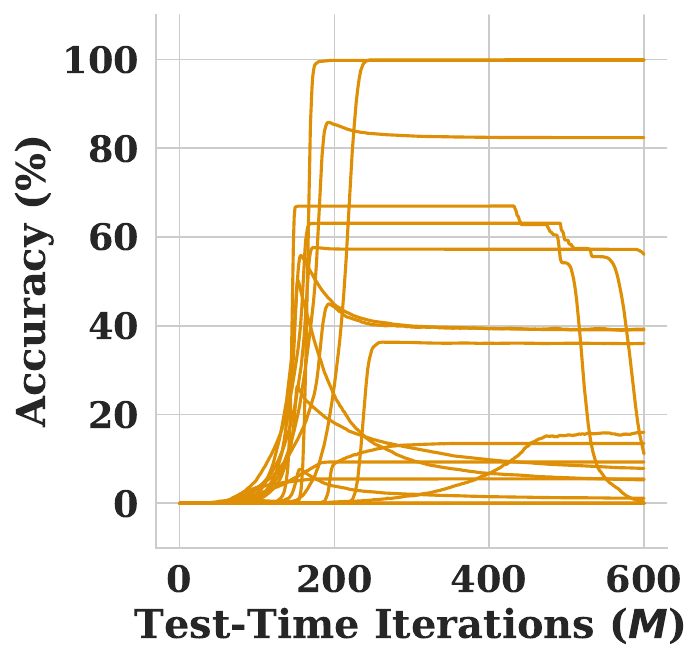}&
        \includegraphics[scale=0.325]{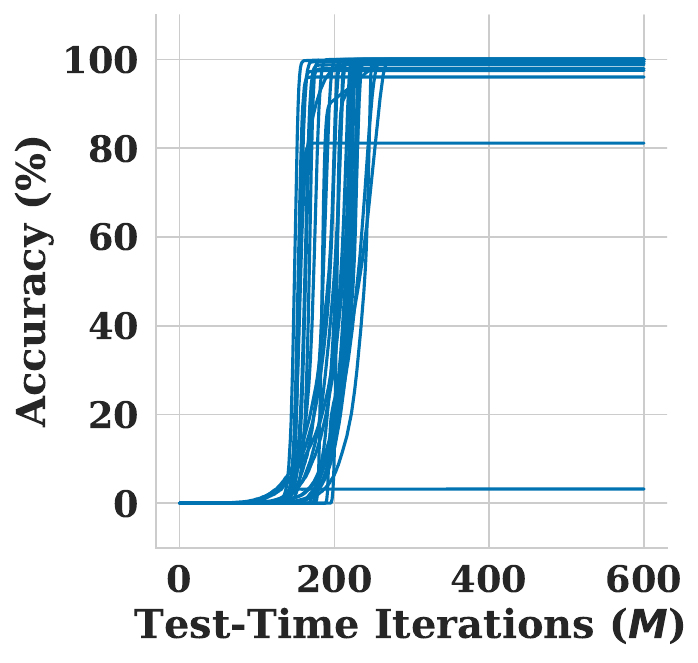}&
        \includegraphics[scale=0.325]{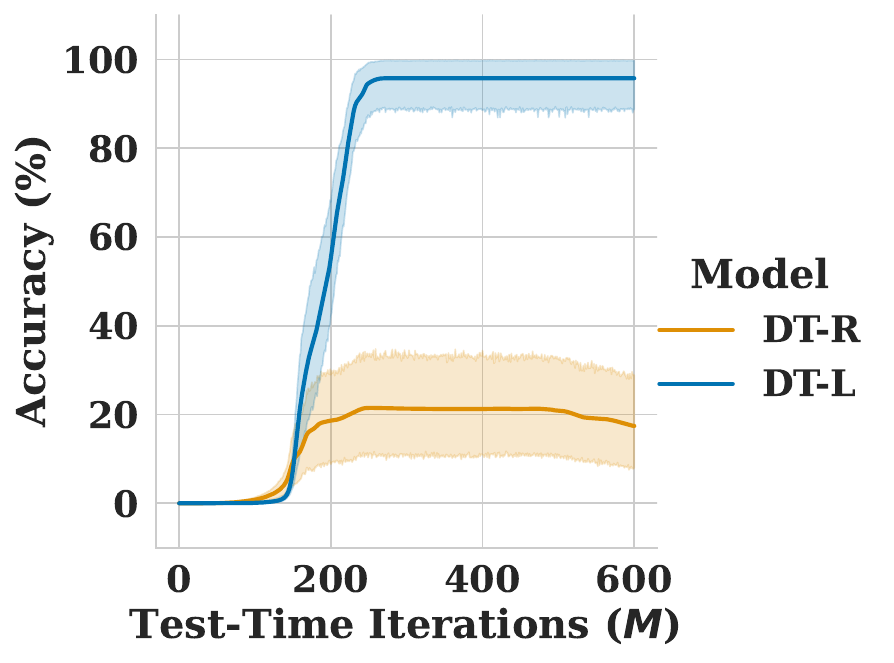}\\
        \DTR{} (runs)&\DTL{} (runs)&Mean
    \end{tabular}
    \captionof{figure}{Comparison between \dtr{} and \dtl{} on the prefix sums
                       problem. Two left plots show the solution accuracy of
                       inference-time runs on 512-bit problems for 30 individual
                       models each. Each line corresponds to the performance of a network trained from scratch with different randomly initial weights. The accuracy is measured on 10\,000 problem instances.  The right plot shows the mean of all 30 for
                       each. Models have a channel width of \(w=32\). Shaded
                       areas show 95\% confidence intervals.}
    \label{fig:prefixsums:runs}
\end{figure}

In our experiments we compare the \DTR{} model to our \DTL{} model. Both were trained on instances consisting of 32 bits for a maximum of 30 iterations, using IPT with \(\alpha=0.5\). Both models have a width (number of channels) of \(w=32\) (more details about architecture can be found in \cref{app:architecture:prefixsums}). We found that for this problem there is little change in performance above $w=32$ for \DTL{}. In \cref{fig:prefixsums:runs} we show the solution accuracy for 30 randomly-initialized models of both \DTR{} and \DTL{} on the 512-bit test dataset versus the number of iterations (\(M\)) to generate those solutions. We measured the performance on 10\,000 randomly generated instances of each problem.  The error in the mean is less than 0.5\% which is approximately the width of the lines in the figure. As can be seen \DTR{} struggles to  consistently find networks that extrapolate to larger problem settings (it only obtains greater than 90\% accuracy on two out of 30 training runs). In contrast \DTL{} only fails to reach above 90\% accuracy on two out of 30 training runs.

To emphasize \DTL{}'s performance on smaller widths, we selected \(w=32\) as
the primary width for comparison. We perform ablations on other aspects of the model in \cref{app:results:ablations}. 

\begin{figure}[t]
    \centering
    \begin{tabular}{ccc}
        \includegraphics[scale=0.325]{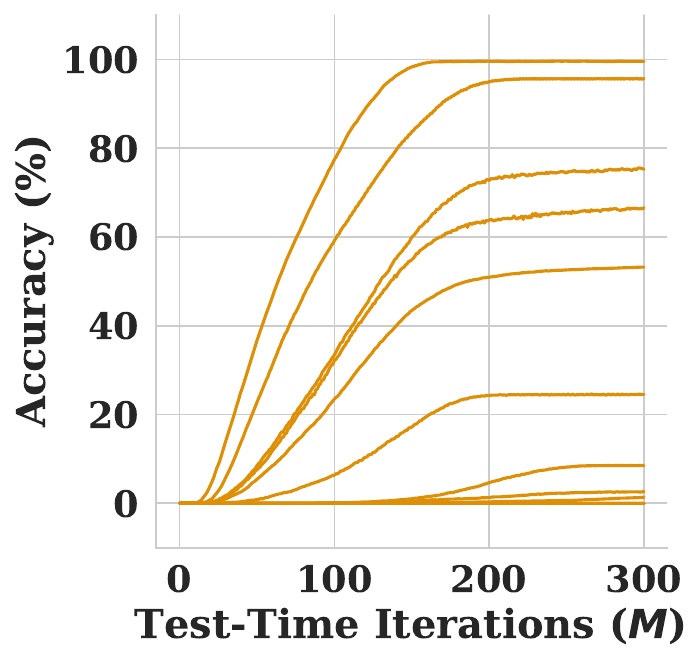}&
        \includegraphics[scale=0.325]{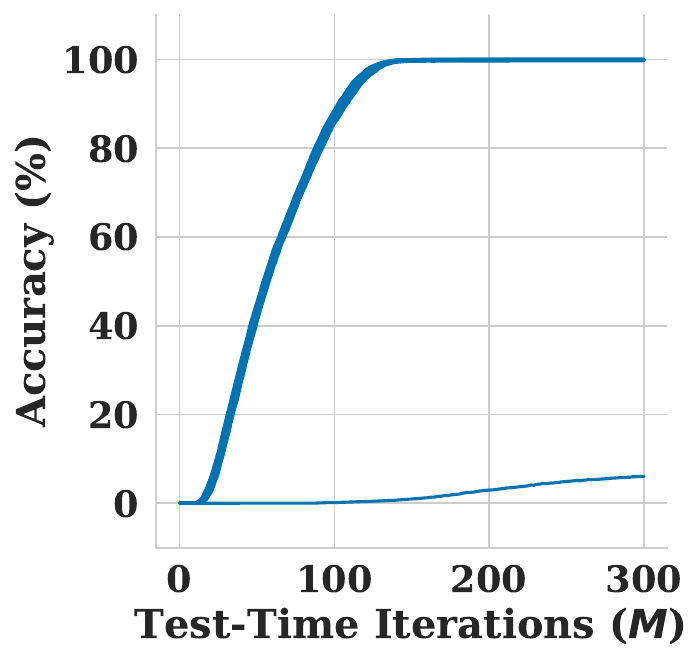}&
        \includegraphics[scale=0.325]{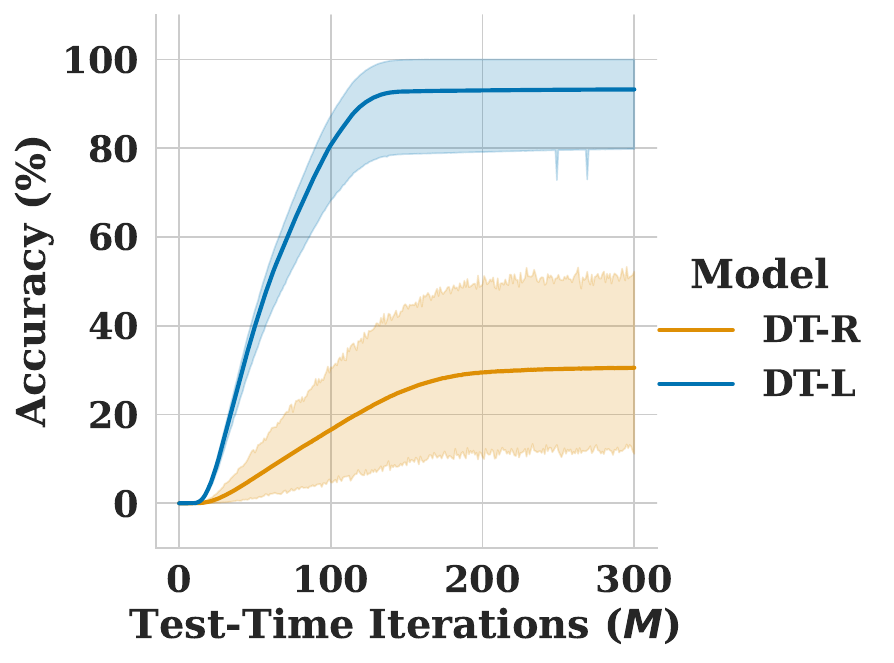}\\
        \DTR{} (runs)&\DTL{} (runs)&Mean
    \end{tabular}
    \captionof{figure}{Comparison between \dtr{} and \dtl{} on the mazes
                       problem for small models. Two left plots show the
                       solution accuracy of inference-time runs on 
                       \(33\times33\) mazes for 14 different models each. The
                       right plot shows the mean of all 14 for each. Models
                       have a channel width of \(w=32\). Shaded areas show
                       95\% confidence intervals.}
    \label{fig:mazes:runs}
\end{figure}

\paragraph{Mazes.}
The mazes problem consists of drawing the correct path for a blank maze, given
a starting point and an ending point. Details about the representation of
mazes is given in~\citeauthor{schwarzschild2021dataset}~\citep{schwarzschild2021dataset}
and~\citet{bansal2022endtoend}.
Our tests are performed on models trained on \(17\times 17\) mazes,
extrapolated to \(33\times 33\) mazes. Results from multiple runs and an
aggregate can be seen in \cref{fig:mazes:runs}.
The maze models defined by \citet{bansal2022endtoend} are trained on \(9\times 9\) mazes.
Attempting to train \DTL{} on mazes of this size resulted in the models often
learning trivial solutions which did not extrapolate to larger mazes.

\begin{figure}[t]
    \centering
    \begin{tabular}{cc}
        \includegraphics[scale=0.325]{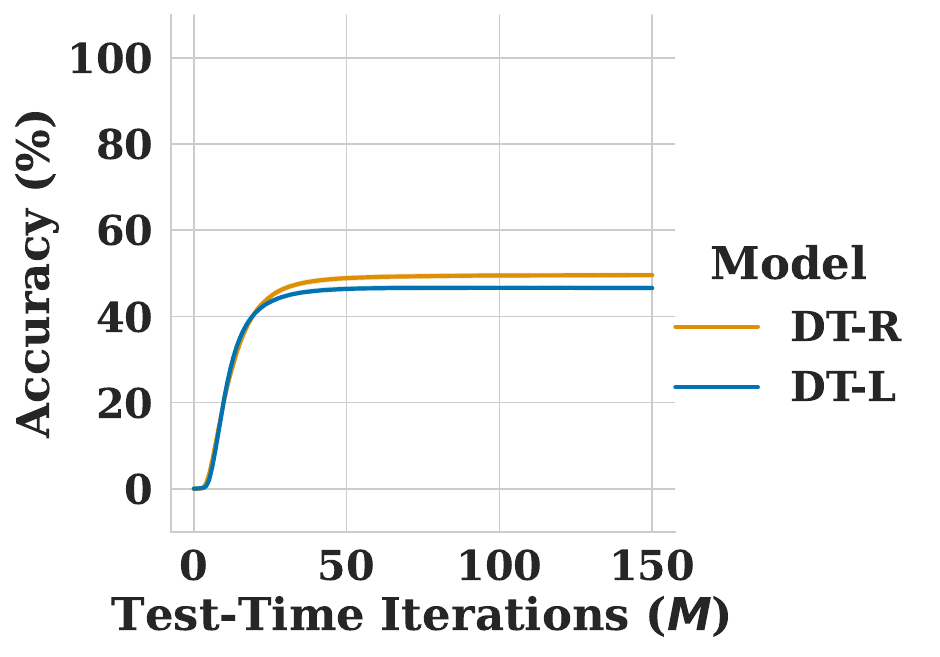}&
        \includegraphics[scale=0.325]{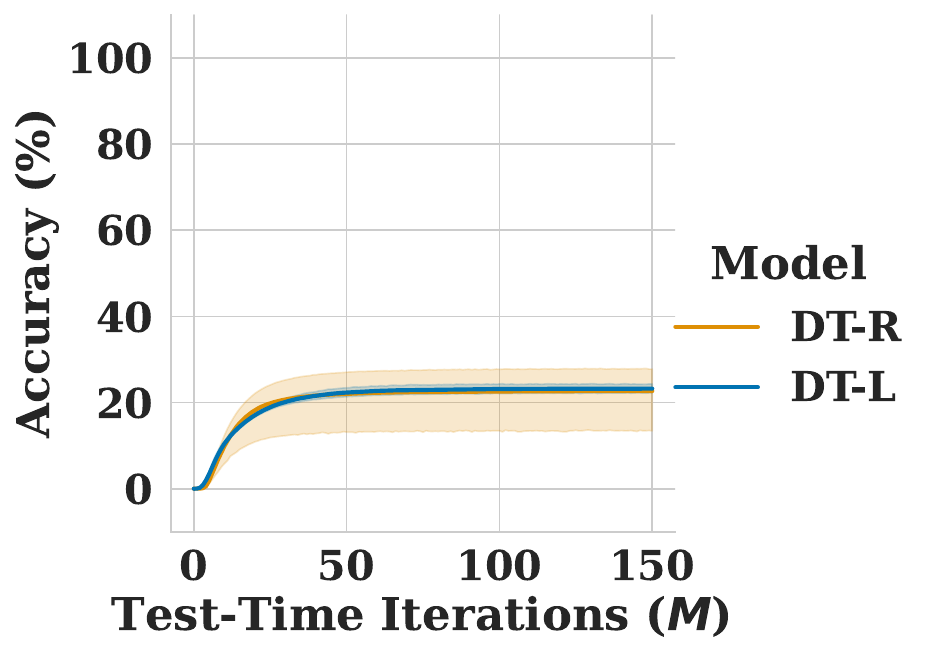}\\
        \(w=64\)&\(w=16\)
    \end{tabular}
    \captionof{figure}{Comparison between \dtr{} and \dtl{} on the chess
                       puzzles problem on models of two different widths,
                       showing the mean accuracy of inference-time runs on
                       problems ranked in increasing difficulty between 1,000,000 and 1,100,000.
                       Aggregate of six models, each trained on the easiest 600,000 problems, where shaded areas show
                       95\% confidence intervals.}
    \label{fig:chesspuzzles:runs}
\end{figure}

\paragraph{Chess Puzzles.}
The chess puzzles problem involves identifying the next best move of the
current board state. Specifically, the model must learn to classify 1s in two
cells representing the piece to move and the location to move to, and 0s
elsewhere. The models are trained on a train/validation split of the easiest
600,000 problems, where difficulty is based on Lichess.org rankings (see
\citet{schwarzschild2021dataset} for more detail).

This problem in particular can be viewed as different to both the prefix sums
and mazes problems in that the difficulty of the problem doesn't come from the
size. All chess puzzles in this dataset are standard \(8\times 8\) chess
boards. 
For all tests on the chess dataset~\citep{schwarzschild2021dataset} we follow the same number of epochs as~\citet{bansal2022endtoend}. Results are shown in \cref{fig:chesspuzzles:runs}. \DTL{} achieves very similar performance to \DTR{}, seemingly hitting the same apparent ceiling on performance discussed by \citet{bansal2022endtoend}. We would like to explore the reasons for this in more detail the future, but we note that at training time the models do not reach 100\% accuracy, so it should not be a surprise that they do not always extrapolate. It is possible that the issue arises from the structure of the data and the way it interacts with the model architecture.

\section{Benchmarking on Traveling Salesperson}\label{section:tsp}

To demonstrate the strength of \DTL, we have formulated the traveling salesperson problem as a differentiable optimization task for the model to solve. This is a significantly more challenging problem than those discussed in \cref{sec:results}.  The problem instance $\bm{x}$ is now the matrices of ``distances'' with additional information.  For arbitrary distance matrices this is known to be NP-hard.  An optimal solution corresponds to a tour (permutation of the cities) such that the sum of distances between neighboring cities is minimized, although we seek only to find a low cost tour.

The solution vector $\bm{y}$ is a binary matrix with 1s corresponding to the edges that are used in the tour.  We constructed the network so that the solution vector would correspond to a feasible solution.  Unfortunately, in doing so involves an intermediate representation where the evaluation of the cost is non-trivial, which made the problem of learning an iterative solution very challenging.  To overcome this, we increased the expressiveness of the scratchpad vectors $\vphi^{(m)}$ to learn orthogonal transforms that it applied to a part of the scratchpad vector.  This considerably improved the performance of the network.  Details of the modification used are given in \cref{app:tsp}.

\subsection{TSP Results}
\label{sec:tsp_results}

In \cref{tab:tsp} we give the mean tour length for randomly generated instances from two class of problems: symmetric random tours and asymmetric random tours.  The \DTL{} models were trained for each problem class on tours of size $n=15$
with \(M=45\).  We show both interpolative results (using a new set of tours on the same size problems) and extrapolative results where we test of problems of size $n=30$ with \(M=120\).  For comparison we provide the length of random tours, the length of tours generated by the greedy nearest neighbor (NN) algorithm, and the length of tours
from a modified version of the NN algorithm which selects the lowest-cost NN tour
out of all starting points instead of starting from a random point (BNN).

The results in \cref{tab:tsp} show \DTL{}'s ability to learn non-trivial
algorithms by performing considerably better than random tours. That the results are worse than the nearest neighbor methods should not be surprising, as in the asymmetric non-Euclidean case these are amongst the best known algorithms.

The models take approximately 8 hours to train on an RTX2080Ti GPU for 80,000 batches of 64 randomly generated TSPs for both symmetric and asymmetric settings with $n=15$ and $M=45$. Testing was performed on M3 Apple Silicon with MPS acceleration and takes 3.59ms per $n=15$ problem instance for $M=45$ and 23ms per $n=30$ problem with $M=120$.

\begin{table}[htbp]
    \caption{Results for TSP runs on symmetric and asymmetric instances.  The results are the mean for 12,800 random instances. For \DTL{}, $M=45$ was used for problem size $n=15$ and $M=120$ for $n=30$, where tour lengths are shown in column 2.  Column 3 (Random Tours) gives the tour length for random tours, while columns 4 and 5 give the tour lengths for
    greedy nearest neighbor tours starting from a random point (NN) or choosing
    the lowest NN tour from all starting points (BNN) respectively.}
    \centering
    \begin{tabular}{lllll}
    \toprule
    Problem & \DTL{} & Random Tours & NN Tours & BNN Tours \\
    \midrule
    Symmetric \(n=15\) & \(3.99\pm 0.006\) & \(7.5\pm 0.01\) & \(2.85\pm 0.005\) & \(2.31\pm 0.005\)\\
    Symmetric \(n=30\) & \(6.02\pm 0.007\) & \(15.0\pm 0.01\) & \(3.50\pm 0.005\) & \(2.72\pm 0.004\)\\
    Asymmetric \(n=15\) & \(4.66\pm 0.008\) & \(7.5\pm 0.01\) & \(2.82\pm 0.006\) & \(2.09\pm 0.004\)\\
    Asymmetric \(n=30\) & \(7.7\pm 0.01\) & \(15.0\pm 0.01\) & \(3.50\pm 0.006\) & \(2.53\pm 0.004\)\\
    \bottomrule
    \end{tabular}
    \label{tab:tsp}
\end{table}


\section{Discussion}
\label{sec:discussion}
\dt{}-style architectures provide a new paradigm for using machine learning for general problem solving.  The key idea is to use a recurrent architecture to find a solution through multiple iterations.  This paper has addressed a major drawback of the published models which is the instability that frequently arises in the iteration steps.  Having addressed these problems not only are we able to obtain networks that much more reliably solve new problems, but we show that we can run much smaller networks with similar and often better performance.  To illustrate the potential of this approach we have tackled a notoriously difficult problem, namely TSP.

\dt{} approaches are attractive because they learn a general problem solving strategy that can be used to solve considerably larger instances of the problem than they were trained on.  Extrapolation to problems outside the training dataset is an area where traditional machine learning struggles.  Clearly, the problems these models extrapolate to are in the same problem class to the training data, but it is noteworthy that the strategies learned for the problem classes we have investigated scale in this way.  Another interesting feature is that the network learns the problem solving strategy through examples (in the case of TSP it is not even shown any low cost solutions).  Admittedly, for TSP we needed to construct a network that outputted tours and to get the quality of results we did we introduced a mechanism for learning orthogonal transformation which we then applied to part of the scratchpad vector.  However, we did not build in any explicit rules for solving the TSP --- so much so that we do not fully understand the algorithm \DTL{} uses to find low cost tours, particularly in the case of asymmetric non-Euclidean TSP.

\paragraph{Broader impact.} Given the ubiquity of iterative algorithms in solving problems, having a mechanism to learn such algorithms opens up a lot of possibilities.  The advantage of using a recurrent mechanism over a feed-forward architecture is that to solve a larger problem we simply needed to run the recurrent loop more often; this is often desirable in the real world because it might be impossible to train on sufficient data that you can guarantee not to need to extrapolate at inference time. Clearly, models of this kind have very many potential applications - both in developing \textit{new} (and potentially improved --- could we learn an algorithm that is significantly faster or more energy efficient than anything currently existing?) algorithms for existing problem classes where we already have solutions, as well as in all areas of predictive modelling where we believe the input-output relationship is best captured through a potentially-deep recurrent function. Inevitably, some of the potential uses also have the potential for misuse.

\paragraph{Limitations.} This work shows theoretically and empirically that it is possible to modify an architecture to learn recurrent functions with convergence guarantees. The problems we solve are relatively simple however.  The performance we obtain, for example, on TSP is far from state-of-the-art, but we believe that the contribution is significant as we are able to find a heuristic algorithm running in $M=O(n\,\log(n))$ steps through trial and error, without any explicit inductive bias towards finding what we might consider to be a sensible algorithm. There are also clearly other potential issues with the approach; learnability is significantly improved over previous attempts, but there are still cases where the model fails to learn. The use of convolutions means that the only way to capture long-term (or large distance) dependencies in the data is through iteration; this is possibly a benefit, but other architectures with different structural biases might work better on certain problems.
We believe, in terms of comparison to NN and BNN heuristics, this
local view of the distance matrix may contribute to the disappointing
appearance of the results.

\paragraph{Outlook.} This work leaves many open directions for future research. It would be interesting to explore different architectures, such as Transformers, and more diverse problem settings. It would also be fascinating to better understand what algorithms are learned by the model under different settings and whether for example the complexity of the learned algorithm can be controlled.

\begin{ack}
JB is funded by a PhD studentship graciously provided by the School of Electronics and Computer Science. JH was supported by the Engineering and Physical Sciences Research Council (EPSRC) International Centre for Spatial Computational Learning [EP/S030069/1]. The authors acknowledge the use of the IRIDIS X High Performance Computing Facility, and the Southampton-Wolfson AI Research Machine (SWARM) GPU cluster generously funded by the Wolfson Foundation, together with the associated support services at the University of Southampton in the completion of this work.
\end{ack}

\bibliographystyle{plainnat}
\bibliography{refs.bib}

\begin{thebibliography}{21}
\providecommand{\natexlab}[1]{#1}
\providecommand{\url}[1]{\texttt{#1}}
\expandafter\ifx\csname urlstyle\endcsname\relax
  \providecommand{\doi}[1]{doi: #1}\else
  \providecommand{\doi}{doi: \begingroup \urlstyle{rm}\Url}\fi

\bibitem[Banach(1922)]{banach}
Stefan Banach.
\newblock {Sur les op\'erations dans les ensembles abstratits et leur applications aux \'equations int\'egrales}.
\newblock \emph{Fundamenat Mathematicae}, 1\penalty0 (3):\penalty0 133 -- 181, 1922.

\bibitem[Bansal et~al.(2022)Bansal, Schwarzschild, Borgnia, Emam, Huang, Goldblum, and Goldstein]{bansal2022endtoend}
Arpit Bansal, Avi Schwarzschild, Eitan Borgnia, Zeyad Emam, Furong Huang, Micah Goldblum, and Tom Goldstein.
\newblock End-to-end algorithm synthesis with recurrent networks: Extrapolation without overthinking.
\newblock In S.~Koyejo, S.~Mohamed, A.~Agarwal, D.~Belgrave, K.~Cho, and A.~Oh, editors, \emph{Advances in Neural Information Processing Systems}, volume~35, pages 20232--20242. Curran Associates, Inc., 2022.
\newblock URL \url{https://proceedings.neurips.cc/paper_files/paper/2022/file/7f70331dbe58ad59d83941dfa7d975aa-Paper-Conference.pdf}.

\bibitem[Ciesielski(2007)]{10.15352/bjma/1240321550}
Krzysztof Ciesielski.
\newblock {On Stefan Banach and some of his results}.
\newblock \emph{Banach Journal of Mathematical Analysis}, 1\penalty0 (1):\penalty0 1 -- 10, 2007.
\newblock \doi{10.15352/bjma/1240321550}.
\newblock URL \url{https://doi.org/10.15352/bjma/1240321550}.

\bibitem[Clevert et~al.(2016)Clevert, Unterthiner, and Hochreiter]{clevert2016elu}
Djork-Arné Clevert, Thomas Unterthiner, and Sepp Hochreiter.
\newblock {Fast and Accurate Deep Network Learning by Exponential Linear Units (ELUs)}, 2016.
\newblock URL \url{https://arxiv.org/abs/1511.07289}.

\bibitem[Graves(2016)]{DBLP:journals/corr/Graves16}
Alex Graves.
\newblock Adaptive computation time for recurrent neural networks.
\newblock \emph{CoRR}, abs/1603.08983, 2016.
\newblock URL \url{http://arxiv.org/abs/1603.08983}.

\bibitem[Graves et~al.(2014)Graves, Wayne, and Danihelka]{graves2014turing}
Alex Graves, Greg Wayne, and Ivo Danihelka.
\newblock Neural turing machines.
\newblock \emph{CoRR}, abs/1410.5401, 2014.
\newblock URL \url{http://arxiv.org/abs/1410.5401}.

\bibitem[Graves et~al.(2016)Graves, Wayne, Reynolds, Harley, Danihelka, Grabska-Barwi{\'{n}}ska, Colmenarejo, Grefenstette, Ramalho, Agapiou, Badia, Hermann, Zwols, Ostrovski, Cain, King, Summerfield, Blunsom, Kavukcuoglu, and Hassabis]{graves2016neural}
Alex Graves, Greg Wayne, Malcolm Reynolds, Tim Harley, Ivo Danihelka, Agnieszka Grabska-Barwi{\'{n}}ska, Sergio~G{\'o}mez Colmenarejo, Edward Grefenstette, Tiago Ramalho, John Agapiou, Adri{\`a}~Puigdom{\`e}nech Badia, Karl~Moritz Hermann, Yori Zwols, Georg Ostrovski, Adam Cain, Helen King, Christopher Summerfield, Phil Blunsom, Koray Kavukcuoglu, and Demis Hassabis.
\newblock Hybrid computing using a neural network with dynamic external memory.
\newblock \emph{Nature}, 538\penalty0 (7626):\penalty0 471--476, Oct 2016.
\newblock ISSN 1476-4687.
\newblock \doi{10.1038/nature20101}.
\newblock URL \url{https://doi.org/10.1038/nature20101}.

\bibitem[Hendrycks and Gimpel(2016)]{hendrycks2016gelu}
Dan Hendrycks and Kevin Gimpel.
\newblock Bridging nonlinearities and stochastic regularizers with gaussian error linear units.
\newblock \emph{CoRR}, abs/1606.08415, 2016.
\newblock URL \url{http://arxiv.org/abs/1606.08415}.

\bibitem[Ho et~al.(2020)Ho, Jain, and Abbeel]{NEURIPS2020_4c5bcfec}
Jonathan Ho, Ajay Jain, and Pieter Abbeel.
\newblock Denoising diffusion probabilistic models.
\newblock In H.~Larochelle, M.~Ranzato, R.~Hadsell, M.F. Balcan, and H.~Lin, editors, \emph{Advances in Neural Information Processing Systems}, volume~33, pages 6840--6851. Curran Associates, Inc., 2020.
\newblock URL \url{https://proceedings.neurips.cc/paper_files/paper/2020/file/4c5bcfec8584af0d967f1ab10179ca4b-Paper.pdf}.

\bibitem[Ioffe and Szegedy(2015)]{sergey2015batchnorm}
Sergey Ioffe and Christian Szegedy.
\newblock Batch normalization: accelerating deep network training by reducing internal covariate shift.
\newblock In \emph{Proceedings of the 32nd International Conference on International Conference on Machine Learning - Volume 37}, ICML'15, page 448–456. JMLR.org, 2015.

\bibitem[Jastrzebski et~al.(2018)Jastrzebski, Arpit, Ballas, Verma, Che, and Bengio]{jastrzebski2018residual}
Stanisław Jastrzebski, Devansh Arpit, Nicolas Ballas, Vikas Verma, Tong Che, and Yoshua Bengio.
\newblock Residual connections encourage iterative inference.
\newblock In \emph{International Conference on Learning Representations}, 2018.
\newblock URL \url{https://openreview.net/forum?id=SJa9iHgAZ}.

\bibitem[Kingma and Ba(2015)]{kingma2015adam}
Diederik Kingma and Jimmy Ba.
\newblock Adam: A method for stochastic optimization.
\newblock In \emph{International Conference on Learning Representations (ICLR)}, San Diega, CA, USA, 2015.

\bibitem[Lu et~al.(2020)Lu, Shin, Su, and Karniadakis]{lu2020dying}
Lu~Lu, Yeonjong Shin, Yanhui Su, and George~Em Karniadakis.
\newblock {Dying ReLU and Initialization: Theory and Numerical Examples}.
\newblock \emph{Communications in Computational Physics}, 28\penalty0 (5):\penalty0 1671–1706, June 2020.
\newblock ISSN 1991-7120.
\newblock \doi{10.4208/cicp.oa-2020-0165}.
\newblock URL \url{http://dx.doi.org/10.4208/cicp.OA-2020-0165}.

\bibitem[Mena et~al.(2018)Mena, Belanger, Linderman, and Snoek]{mena2018gumbel}
Gonzalo Mena, David Belanger, Scott Linderman, and Jasper Snoek.
\newblock {Learning Latent Permutations with Gumbel-Sinkhorn Networks}.
\newblock In \emph{International Conference on Learning Representations}, 2018.
\newblock URL \url{https://openreview.net/forum?id=Byt3oJ-0W}.

\bibitem[Miyato et~al.(2018)Miyato, Kataoka, Koyama, and Yoshida]{miyato2018spectral}
Takeru Miyato, Toshiki Kataoka, Masanori Koyama, and Yuichi Yoshida.
\newblock {Spectral Normalization for Generative Adversarial Networks}.
\newblock In \emph{International Conference on Learning Representations}, 2018.
\newblock URL \url{https://openreview.net/forum?id=B1QRgziT-}.

\bibitem[Paszke et~al.(2019)Paszke, Gross, Massa, Lerer, Bradbury, Chanan, Killeen, Lin, Gimelshein, Antiga, Desmaison, Kopf, Yang, DeVito, Raison, Tejani, Chilamkurthy, Steiner, Fang, Bai, and Chintala]{paszke2019torch}
Adam Paszke, Sam Gross, Francisco Massa, Adam Lerer, James Bradbury, Gregory Chanan, Trevor Killeen, Zeming Lin, Natalia Gimelshein, Luca Antiga, Alban Desmaison, Andreas Kopf, Edward Yang, Zachary DeVito, Martin Raison, Alykhan Tejani, Sasank Chilamkurthy, Benoit Steiner, Lu~Fang, Junjie Bai, and Soumith Chintala.
\newblock Pytorch: An imperative style, high-performance deep learning library.
\newblock In H.~Wallach, H.~Larochelle, A.~Beygelzimer, F.~d\textquotesingle Alch\'{e}-Buc, E.~Fox, and R.~Garnett, editors, \emph{Advances in Neural Information Processing Systems}, volume~32. Curran Associates, Inc., 2019.
\newblock URL \url{https://proceedings.neurips.cc/paper_files/paper/2019/file/bdbca288fee7f92f2bfa9f7012727740-Paper.pdf}.

\bibitem[Schwarzschild et~al.(2021{\natexlab{a}})Schwarzschild, Borgnia, Gupta, Bansal, Emam, Huang, Goldblum, and Goldstein]{schwarzschild2021dataset}
Avi Schwarzschild, Eitan Borgnia, Arjun Gupta, Arpit Bansal, Zeyad Emam, Furong Huang, Micah Goldblum, and Tom Goldstein.
\newblock {Datasets for Studying Generalization from Easy to Hard Examples}.
\newblock \emph{CoRR}, abs/2108.06011, 2021{\natexlab{a}}.
\newblock URL \url{https://arxiv.org/abs/2108.06011}.

\bibitem[Schwarzschild et~al.(2021{\natexlab{b}})Schwarzschild, Borgnia, Gupta, Huang, Vishkin, Goldblum, and Goldstein]{schwarzschild2021algorithm}
Avi Schwarzschild, Eitan Borgnia, Arjun Gupta, Furong Huang, Uzi Vishkin, Micah Goldblum, and Tom Goldstein.
\newblock Can you learn an algorithm? generalizing from easy to hard problems with recurrent networks.
\newblock In M.~Ranzato, A.~Beygelzimer, Y.~Dauphin, P.S. Liang, and J.~Wortman Vaughan, editors, \emph{Advances in Neural Information Processing Systems}, volume~34, pages 6695--6706. Curran Associates, Inc., 2021{\natexlab{b}}.
\newblock URL \url{https://proceedings.neurips.cc/paper_files/paper/2021/file/3501672ebc68a5524629080e3ef60aef-Paper.pdf}.

\bibitem[Trottier et~al.(2016)Trottier, Gigu{\`{e}}re, and Chaib{-}draa]{trottier2016pelu}
Ludovic Trottier, Philippe Gigu{\`{e}}re, and Brahim Chaib{-}draa.
\newblock {Parametric Exponential Linear Unit for Deep Convolutional Neural Networks}.
\newblock \emph{CoRR}, abs/1605.09332, 2016.
\newblock URL \url{http://arxiv.org/abs/1605.09332}.

\bibitem[Yoshida and Miyato(2017)]{yoshida2017spectral}
Yuichi Yoshida and Takeru Miyato.
\newblock {Spectral Norm Regularization for Improving the Generalizability of Deep Learning}, 2017.
\newblock URL \url{https://arxiv.org/abs/1705.10941}.

\bibitem[Zhang et~al.(2019)Zhang, Hare, and Prugel-Bennett]{NEURIPS2019_6e79ed05}
Yan Zhang, Jonathon Hare, and Adam Prugel-Bennett.
\newblock Deep set prediction networks.
\newblock In H.~Wallach, H.~Larochelle, A.~Beygelzimer, F.~d\textquotesingle Alch\'{e}-Buc, E.~Fox, and R.~Garnett, editors, \emph{Advances in Neural Information Processing Systems}, volume~32. Curran Associates, Inc., 2019.
\newblock URL \url{https://proceedings.neurips.cc/paper_files/paper/2019/file/6e79ed05baec2754e25b4eac73a332d2-Paper.pdf}.

\end{thebibliography}

\newpage
\appendix
\setcounter{table}{0}
\renewcommand{\thetable}{\Alph{section}\arabic{table}}

\setcounter{figure}{0}
\renewcommand{\thefigure}{\Alph{section}\arabic{figure}}

\setcounter{equation}{0}
\renewcommand{\theequation}{\Alph{section}\arabic{equation}}

\section{On Being Lipschitz}
\label{app:proofs}

As discussed in the main paper one of its major contribution is the observation that by making $\networkG(\cdot,\bm{x})$ \(K\)-Lipschitz with $K<1$, then the iterative mapping defined in \cref{eq:iteration} is guaranteed to converge to a unique solution as a consequence of the Banach fixed-point theorem~\cite{banach}.  To engineer $\networkG(\cdot,\bm{x})$ to be \(K\)-Lipschitz we rely on a few well known properties of functions.  If $\mathcal{C}:X\rightarrow Z$ is a composition of two Lipschitz mappings $\mathcal{A}:X\rightarrow Y$ and $\mathcal{B}: Y \rightarrow Z$ with Lipschitz constant $K_A$ and $K_B$ then
\begin{align}
    \norm{\mathcal{C}(\bm{x})-\mathcal{C}(\bm{y})}
    = \norm{\mathcal{B}(\mathcal{A}(\bm{x})) - \mathcal{B}(\mathcal{A}(\bm{y}))} \leq K_B\, \norm{\mathcal{A}(\bm{x})-\mathcal{A}(\bm{y})} \leq K_A\,K_B \norm{\bm{x} - \bm{y}}.
\end{align}
Thus a sufficient condition to ensure $\mathcal{C}$ is \(K\)-Lipschitz with $K<1$ is that $K_A, \, K_B < 1$.  This trivially generalises to any number of compositions.

In the network $\networkG(\vphi, \bm{x})$ the input vector $\bm{x}$ is added to $\vphi$ through a convolution.  However, adding an offset vector to a mapping does not affect the Lipschitz property as if $\mathcal{A}$ is K-Lipschitz and $\bm{c}$ is a constant vector, then the mapping $\mathcal{B}(\vx) = \mathcal{A}(\vx) + \vc$ satisfies
\begin{align}
    \norm{\mathcal{B}(\bm{x}) - \mathcal{B}(\bm{y})}
    = \norm{ \mathcal{A}(\bm{x}) - \mathcal{A}(\bm{y})}
    \leq K\, \norm{\bm{x}-\bm{y}}.
\end{align}
Since during every iteration the input vector $\bm{x}$ and the convolution filters remain unchanged, the addition of the input $\vx$ (which the \dtr{} authors termed the recall) does not change the Lipschitz property of the network.

Finally, to ensure the Lipschitz behaviour of the convolutions we note that convolutions are equivalent to applying a matrix, $\mat{C}$, to some vector, $\vv$.  For a matrix norm $\norm{\mat{C}}$ that is compatible with a vector norm $\norm{\bm{x}}$,
\begin{align}
    \norm{ \mat{C}\,\bm{v}} \leq \norm{\mat{C}} \, \norm{\bm{x}}\;,
\end{align}
where there exists a vector $\bm{x}$ where the equality conditions holds.  If $N=\norm{C}$ then, through the linearity of norms, $\mat{D} = (K/N)\, \mat{C}$ will have norm \(K/N\) so that the linear mapping $\mathcal{D}(\vx) = \mat{D} \,\vx$ is \(K\)-Lipschitz since
\begin{align}
    \norm{\mathcal{D}(\bm{x}) - \mathcal{D}(\bm{y})}
    = \norm{\mat{D}(\bm{x}-\bm{y})} 
    = \frac{K}{N} \norm{\mat{C}(\bm{x}-\bm{y})}
    \leq \frac{K}{N} \norm{\mat{C}} \, \norm{\bm{x}-\bm{y}}
    = K\,\norm{\bm{x}-\bm{y}}.
\end{align}
This would be true for any normed vector space with an appropriate compatible matrix norm.  In this paper we have used the $\ell_2$ norm where the compatible matrix norm is the spectral norm (i.e. the largest singular value).

To allow the algorithm being run by network~\networkG to accumulate enough information to find a good solution we want the Lipschitz constant, $K$, of network~\networkG to be as close to 1 as possible.  If the network can find a good solution rapidly then it can find a scratchpad vector $\vphi$ in a region where the effective Lipschitz constant is less than 1.  Indeed, we observed that despite \dtr{} most often having Lipschitz constant greater than 1, nevertheless in the few runs where a successful network was learned, the growth in the solution $\norm{\vphi^{(m+1)}}/\norm{\vphi^{(m)}}$ could be less than 1.  Being on the edge of the stable region seems to be helpful for the network to find good solutions to hard problems.  Consequently, in engineering the network~\networkG we attempt to make the Lipschitz constant for most mapping layers as close to 1 as possible.

\section{Architecture}\label{app:architecture}

In this section we describe the architecture of \DT{} \cite{schwarzschild2021dataset}, \DTR{} \cite{bansal2022endtoend} and our network \DTL. For \DT{} and \DTR{} the architecture is identical to that described in the original paper.  We include a description here to make the comparison with \DTL{} clearer.  The number of channels of the input, $\bm{x}$, varies depending on the problem class.  It is 1 for prefix sums, 3 for mazes, 12 for chess and 3 for TSP.  Following \DT{} and \DTR{}, the output vector, $\bm{y}$, has two channels.  During training this is used to compute a categorical cross-entropy loss, while at inference time this is put into a max function to obtain a binary output.

In all models the convolutions are either 3 (for prefix sum) or $3\times3$ with stride 1 and padding 1.  In the figures below the boxes with rounded edges represent convolutions. 

\paragraph{\dt.} In \cref{fig:dt-model} we show the \dt{} (\DT) model.  Note in this case the input vector $\bm{x}$ is not given to the network at each iteration of \networkG.

\begin{figure}[H]
    \centering
    \resizebox{\textwidth}{!}{\input{Resources/Images/dt_model.tikz}}
    \caption{The \DT{} architecture of \citet{schwarzschild2021algorithm}. Arrows denote flow from one function to the next with recurrent iterations are denoted by the dashed arrow. Single-lined rounded rectangles denote regular convolutions. Activation functions in square boxes are ReLU.}
    \label{fig:dt-model}
\end{figure}

\paragraph{\dtr.} In \cref{fig:dtr-model} we show the architecture for the \dtr{} (\DTR) model. In \DTR{} the network is fed $\vx$ every iteration and it is concatenated to the recurrent connection.

\begin{figure}[H]
    \centering
    \resizebox{\textwidth}{!}{\input{Resources/Images/dtr_model.tikz}}
    \caption{The \DTR{} architecture of \citet{bansal2022endtoend}. Arrows denote flow from one function to the next with recurrent iterations are denoted by the dashed arrow. Single-lined rounded rectangles denote regular convolutions. The recall $\vx$ is concatenated ($\|$) to the recurrent input. Activation functions in square boxes are ReLU by default, but we experiment with ELU in \cref{app:results:ablations}.}
    \label{fig:dtr-model}
\end{figure}

\begin{figure}[H]
    \centering
    \resizebox{\textwidth}{!}{\input{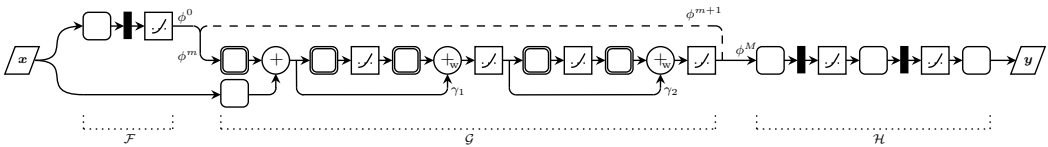}}
    \caption{Our basic \DTL{} architecture. Arrows denote flow from one function to the next with recurrent iterations are denoted by the dashed arrow. Double-lined rounded rectangles denote convolutions with the Lipschitz constraint applied. Single-lined rounded rectangles are regular convolutions. Solid black rectangles denote batch-normalization operations. Activation functions are ELU by default, but we experiment with ReLU in \cref{app:results:ablations}. Plus symbols with a small $\mathrm{w}$ indicate weighted summation, with the residual weighted by $1-\gamma$ and main branch by $\gamma$ (c.f. \cref{eqn:parametric_residual}). Rather than concatenating the recall $\vx$ into the recurrent input we pre-process it with a convolution and then sum with a convolved version of the recurrent input. }
    \label{fig:dtl-model}
\end{figure}

\paragraph{\dtl.} The architecture of our model, \dtl{} (DTL) is shown in \cref{fig:dtl-model}.  The convolutions with spectral norms are shown in bold.  Following common practice when working with models with skip connections we have also added batch norms (shown as solid black rectangles).  These ensure that the tensors going through the network are normalized ensuring that their means are around zero where the non-linearity in the activation functions are strongest.  Empirically this lead to improved performance as shown by ablation studies.

\subsection{TSP Model}
\label{app:tsp}

Recall that in TSP the input vector $\bm{x}$ corresponds to the matrices of distances\footnote{Technically these are pseudo-distances as although they are positive they are not necessarily symmetric and are not guaranteed to satisfy the triangular inequality.}. The solution $\bm{y}$ is a binary matrix showing which edges we used in the tour.  Thus, the cost of the full tour is $\operatorname{sum}(\bm{x}\odot \bm{y})$, where $\odot$ denotes element-wise multiplication. This is just the sum the set of edge distances that are used in the tour.  This is schematically shown below.
\begin{center}
    \includegraphics[width=14cm]{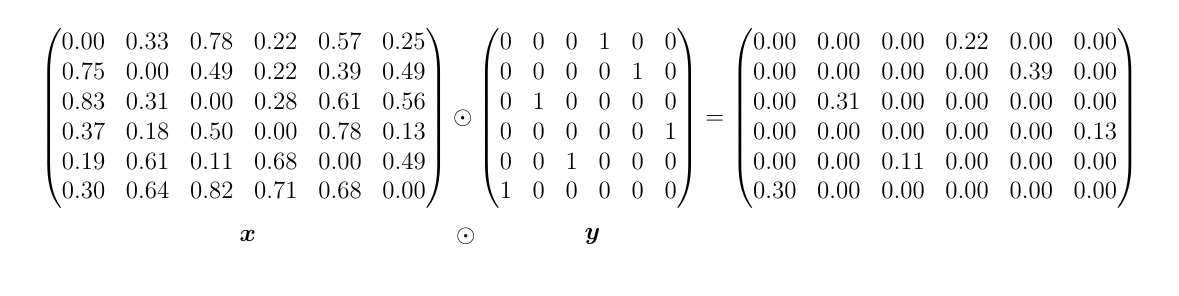}
\end{center}
The edge matrix $\bm{y}$ is a matrix with a 1 in each row and each column and zero elsewhere.  That is, it is a permutation matrix.  Unfortunately, not all permutation matrices correspond to legal tours.  The only permutations that are allowed correspond to irreducible permutation (i.e. those consisting of a single cycle of length $n$, where $n$ is the number of cities)---these are cyclic permutations of order $n$.  This, arises because there are $(n-1)!$ tours (since the starting city is arbitrary) whereas there are $n!$ permutation matrices.  The illegal permutations consist of multiple cycles (e.g. for a four city problem we might have $1\leftrightarrow 2$ and $3\leftrightarrow 4$.).  It is however challenging to generate irreducible permutations.  To solve this problem we use the fact that the set of permutations of order $n$ form an equivalence class $[n]$ and for any permutation $\bm{q} \in S_n$ (where $S_n$ is the group of all permutations of $n$ objects---known as the symmetric group) then if $\bm{\pi} \in [n]$ the conjugate $\bm{q}\,\bm{\pi}\,\bm{q}^{-1} \in [n]$.  This is a well known result in group theory, but it means that if we start with an irreducible representation $\bm{\pi}$ (e.g. the tour $1\rightarrow 2 \rightarrow 3 \rightarrow \cdots \rightarrow n \rightarrow 1$) then for any permutation matrix $\bm{q}$ the product $\bm{q}\,\bm{\pi}\,\bm{q}^{-1}$ represents a legal tour.  Thus we set up the networks to generate a permutation matrix $\bm{q}$.
This used a Gumbel-Sinkhorn network \cite{mena2018gumbel} as part of network \networkH.  Occasionally this would generate a solution with fractional edges.  To avoid this we added a term to the loss function that punished solutions with fractional edges.

Although this procedure guarantees that the solutions correspond to feasible tours, the quality of tours obtained when we trained the full network is poor.  The problem is that the relationship between the permutations $\bm{q}$ and the matrices of edges visited $\bm{y} = \bm{q}\,\bm{\pi}\,\bm{q}^{-1}$ is difficult to learn.  To address this problem in the scratch-pad vector we learned three groups of channels $\vphi = (\vphi_1, \vphi_2, \bm{m})$.  The last part $\bm{m}$ is a single channel which we treat as a matrix from which we compute an orthogonal transformation.  To achieve this we perform SVD to obtain $\bm{u}\bm{s}\bm{v}^\tr$.  We discard $\bm{s}$ to obtain an orthogonal matrix $\bm{w} = \bm{u}\bm{v}^\tr$.  We use this to transform $\vphi_2$.  This reordering significantly improves the quality of the tours we learn.  We attribute this to providing the network \networkG with the ability to understand how altering $\bm{q}$ will change $\bm{y} = \bm{q}\,\bm{\pi}\,\bm{q}^{-1}$ and hence the cost of a tour.  We used a differentiable version of SVD from the PyTorch library \citep{paszke2019torch}. Very occasionally this would fail, in which case we terminated the training and start again.  Although rare this was the one source of failure of the network.

For asymmetric tours we obtained improved results by making the input $\bm{x}$ have three channels.  The first being the distance matrix and the second the transpose of the distance matrix and the third a matrix that have values of 1 in the upper triangle, 0 along the diagonal and -1 in the low triangle.  For convenience the same input was used for symmetric instances, but obviously the transposed distance matrix is equal to the distance matrix.  In addition, the scratch-pad vector learnt two matrices $\vphi = (\vphi_1, \vphi_2, \bm{m}_1, \bm{m}_2)$ from which we constructed two orthogonal matrices $\bm{w}_i = \bm{u}_i\bm{v}_i^\tr$ which we used to apply both column-wise and row-wise transformations \(\vw_1\vphi_2\vw_2^\tr\).

With these modifications we then trained \DTL{} on randomly generated instances
of TSP using a loss function based on the mean cost of a single edge in the
selected tour. The distances were normalized so the maximum edge is 1. The loss function can be stated as
\begin{equation}
    \mathcal{L}_{\mathrm{tour}}=\dfrac{1}{N}\sum_{ij}\left[\left(
    x_{ij}-1\right)\cdot y_{ij}^{\xi}\right]
\end{equation}
where \(\xi\geq 1\) is a hyperparameter that pushes values not equal to 1 closer to zero, thus preventing solutions with fractional edges.

\subsection*{Prefix Sums Models}
\label{app:architecture:prefixsums}

For different widths \(w\), we scale the channels of the convolutional
layers in \networkH (output module) as follows:
\renewcommand{\arraystretch}{1.5}
\begin{center}\begin{tabular}{lrr}
    \toprule\textbf{Layer}&\textbf{Input Channels}&\textbf{Output Channels}\\ \midrule
    Convolution 1&\(w\)&\(w\)\\
    Convolution 2&\(w\)&\(\max\left(2,\left\lfloor w/2\right\rfloor
    \right)\)\\
    Convolution 3&\(\max\left(2,\left\lfloor w/2\right\rfloor\right)\)&
    2\\
    \bottomrule
\end{tabular}\end{center}
\renewcommand{\arraystretch}{1}
This ensures consistency with the \(w=400\) prefix sums models in
\citep{bansal2022endtoend}. For \(w=32\), this results in output channels of
32, 16, and 2.

\subsubsection*{Mazes Models}
\label{app:architecture:mazes}

For different widths \(w\), we scale the channels of the convolutional layers
in \networkH (output module) as follows:
\renewcommand{\arraystretch}{1.5}
\begin{center}\begin{tabular}{lrr}
    \toprule\textbf{Layer}&\textbf{Input Channels}&\textbf{Output Channels}\\ \midrule
    Convolution 1&\(w\)&\(\max\left(2,4^{-1}w\right)\)\\
    Convolution 2&\(\max\left(2,4^{-1}w\right)\)&\(\max\left(
    2,4^{-2}w\right)\)\\
    Convolution 3&\(\max\left(2,4^{-2}w\right)\)&2\\
    \bottomrule
\end{tabular}\end{center}
\renewcommand{\arraystretch}{1}

\subsubsection*{Chess Puzzles Models}
\label{app:architecture:chesspuzzles}

For different widths \(w\), we scale the channels of the convolutional layers
in \networkH (output module) as follows:
\renewcommand{\arraystretch}{1.5}
\begin{center}\begin{tabular}{lrr}
    \toprule\textbf{Layer}&\textbf{Input Channels}&\textbf{Output Channels}\\ \midrule
    Convolution 1&\(w\)&\(\max\left(2,16^{-1}w\right)\)\\
    Convolution 2&\(\max\left(2,16^{-1}w\right)\)&2\\
    Convolution 3&2&2\\
    \bottomrule
\end{tabular}\end{center}
\renewcommand{\arraystretch}{1}

\section{Training}\label{app:training}

Unless specified otherwise:
\begin{itemize}
    \item All models use the Adam optimizer \citep{kingma2015adam} with
          \begin{itemize}
              \item a learning rate of 0.001,
              \item \(\beta_1=0.9\), \(\beta_2=0.999\),
              \item weight decay set to 0.0002 and only applied to
                    unconstrained convolutional weights;
          \end{itemize}
    \item incremental progress training with \(\alpha=0.5\);
    \item exponential warmup with a warmup period of 3;
    \item a multi-step learning rate scheduler where milestones are calculated
          as a \(8:4:2:1\) ratio of the total number of epochs, with learning
          rates multiplied by 0.1 at each milestone.
\end{itemize}
As an example, prefix-sums-solving models are trained to 150 epochs. With the
given ratio this produces milestones of 80, 120, and 140.

For all models, the end-of-epoch state resulting in the best validation score
(accuracy for non-TSP problems, loss for TSP) became the final result of
training. This is standard practice in training deep learning models.

\subsection{Prefix Sums}
Prefix-sums-solving models were trained on 32-bit prefix
sums from the EasyToHard dataset \citep{schwarzschild2021dataset} for 150
epochs, shuffled and split into 80\% training samples, 20\% validation samples.
Each batch contained 500 samples.

Trained models for ablation studies instead used milestones at a \(4:2:1\)
ratio of 150 epochs instead.

Models were trained with \(M=30\).

\subsection{Mazes}
Maze-solving models were trained on \(17\times 17\) mazes from the EasyToHard
dataset~\citep{schwarzschild2021dataset} for 50 epochs, shuffled and split
into 80\% training samples, 20\% validation samples. Each batch contained 50
samples.

Models were trained with \(M=30\).

\subsection{Chess Puzzles}
Chess puzzles models were trained on the easiest 600,000 chess instances from the EasyToHard dataset~\citep{schwarzschild2021dataset} for 120 epochs, with batch sizes of 300
problem instances. Learning rate scheduling followed the same
\(8:4:2:1\) milestone ratio, with a multiplier of 0.1. The dataset was
shuffled and split into 80\% training samples, 20\% validation samples. Each
batch contained 300 samples.

Models were trained with \(M=30\).

\subsection{Traveling Salesperson}
Models trained to produce tours for TSP instead use stochastic gradient
descent (SGD) with a learning rate of 0.001, Nesterov momentum of 0.9,
and a weight decay of 0.0002 (applied only to unconstrained convolutional
weights).

The model trained to perform symmetric TSP used IPT with \(\alpha=0.5\),
whereas the model trained to perform asymmetric TSP did not use IPT.

Since there is no dataset, we chose 1,000 samples to be an appropriate size
for one epoch. With 80\% training samples and 20\% validation samples, this
produces 800 training batches per epoch and 200 validation samples per epoch.
We generate 64 grids of distances per batch.

\section{Additional Results and Analysis}

\subsection{\DTR{} Stability}
\label{app:results:dtstability}
\begin{figure}[H]
    \centering
    \begin{tabular}{ccc}
        \includegraphics[scale=0.25]{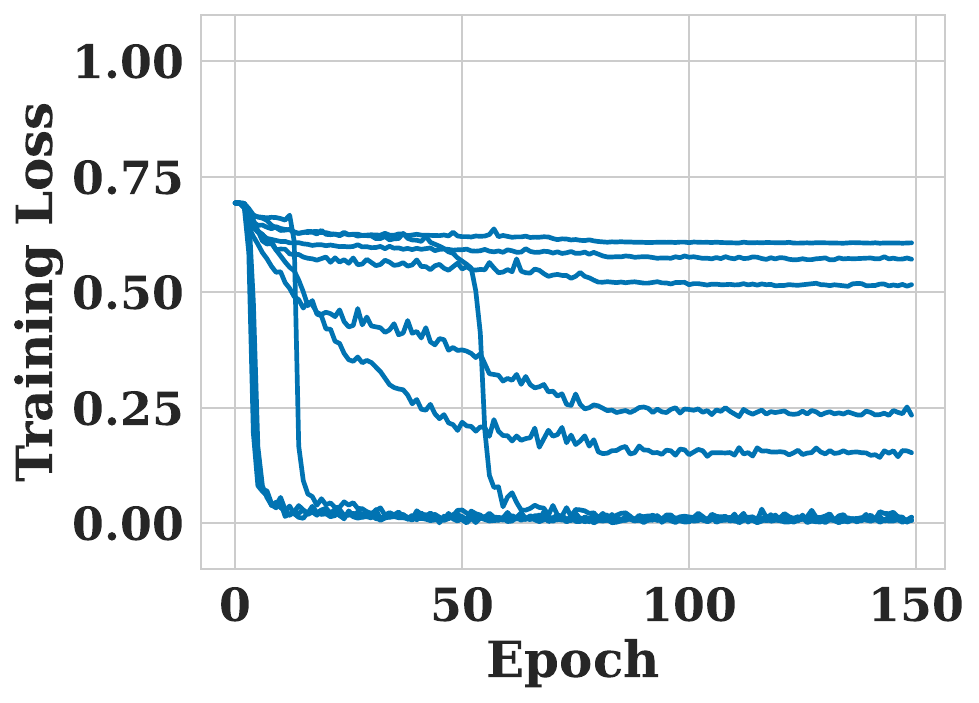}&
        \includegraphics[scale=0.25]{Resources/Plots/Stability/w16.pdf}&
        \includegraphics[scale=0.25]{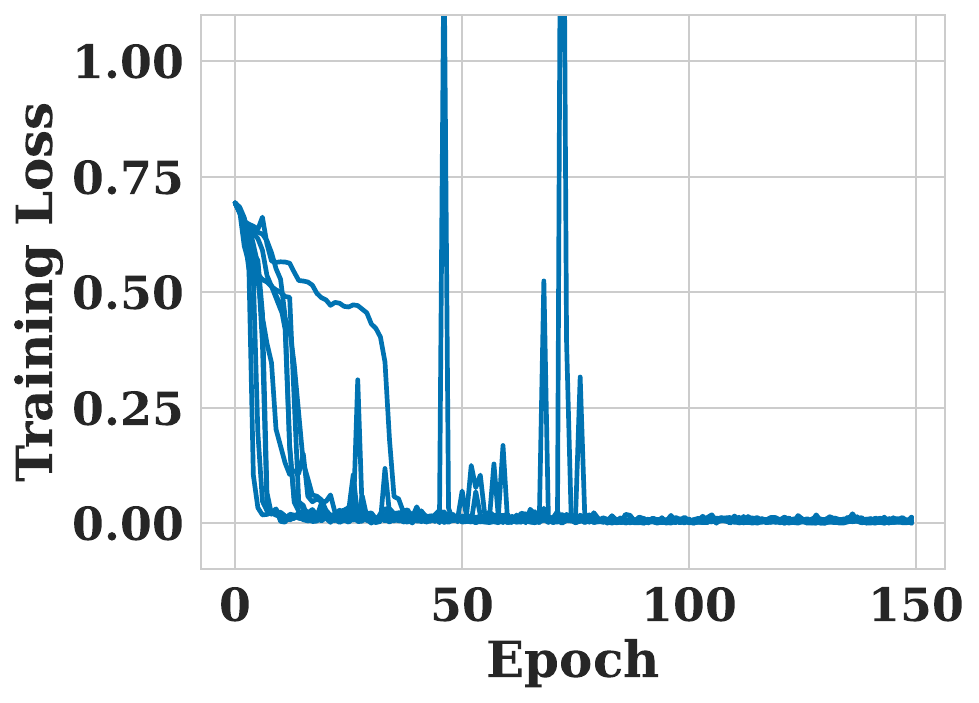}\\
        \(w=8\)&\(w=16\)&\(w=32\)\\\\
        \includegraphics[scale=0.25]{Resources/Plots/Stability/w64.pdf}&
        \includegraphics[scale=0.25]{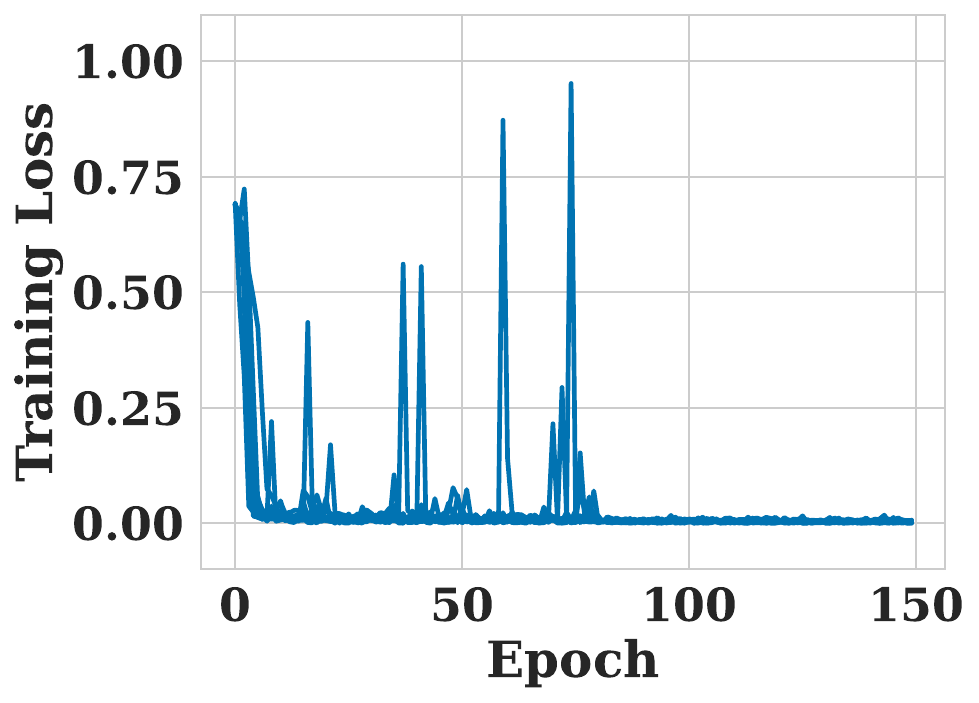}&
        \includegraphics[scale=0.25]{Resources/Plots/Stability/w256.pdf}\\
        \(w=64\)&\(w=128\)&\(w=256\)\\\\&
        \includegraphics[scale=0.25]{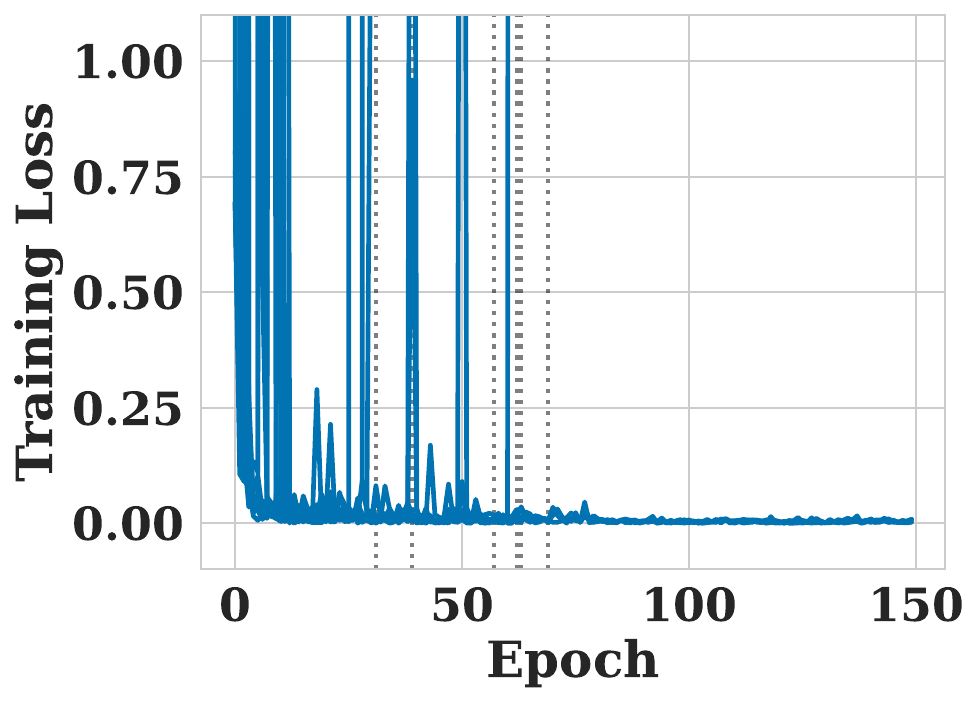}\\
        &\(w=512\)&
    \end{tabular}
    \captionof{figure}{Extended results of \cref{fig:dtanalysis:stability}.
                       Mean training (cross-entropy) loss at each epoch for
                       prefix-sums-solving models of varying width \(w\). Each
                       curve is measured from a different random
                       initialization of the model throughout training, for
                       10 models of each width. Dotted vertical lines indicate
                       loss becoming NaN or infinite, where the model does not
                       recover.}
    \label{fig:dtanalysis:stabilityext}
\end{figure}

To show the training instability in more widths \(w\) for the prefix sums
\DTR{} models, \cref{fig:dtanalysis:stabilityext} shows extended results from
\cref{fig:dtanalysis:stability}.

\subsection{Ablation Studies}
\label{app:results:ablations}

We have performed ablation studies on prefix-sums-solving models of
width \(w=32\) and using IPT with \(\alpha=1.0\). Instead of the
\(8:4:2:1\) milestone ratio (\cref{app:training}), these models use a
\(4:2:1\) milestone ratio, meaning the multiplier 0.1 is only applied twice
instead of three times.

\begin{figure}[H]
    \centering
    \begin{tabular}{cc}
        \includegraphics[scale=0.37]{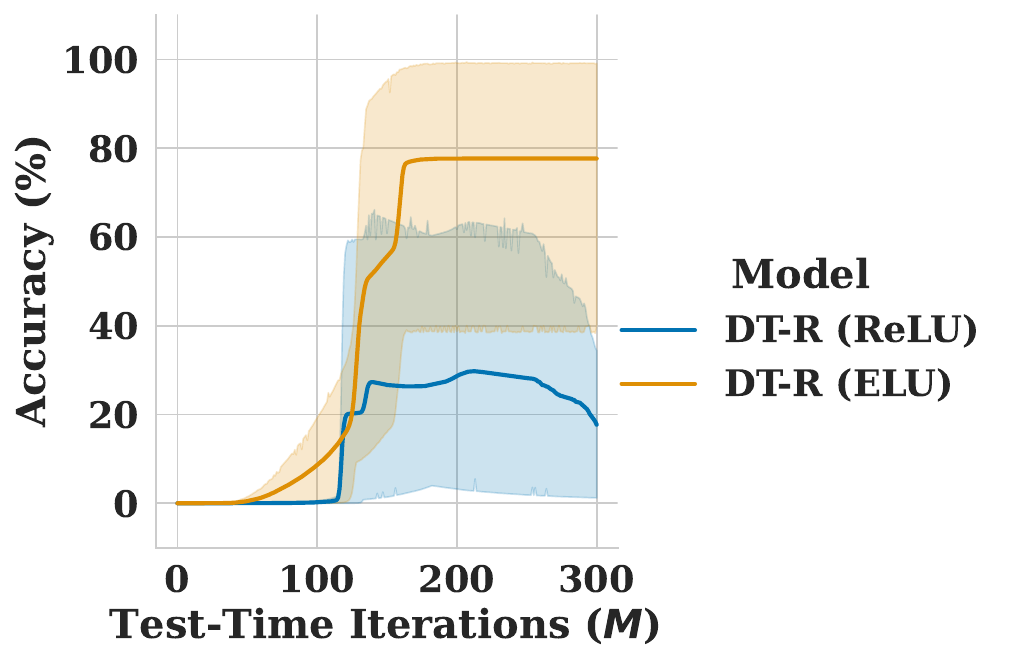}&
        \includegraphics[scale=0.37]{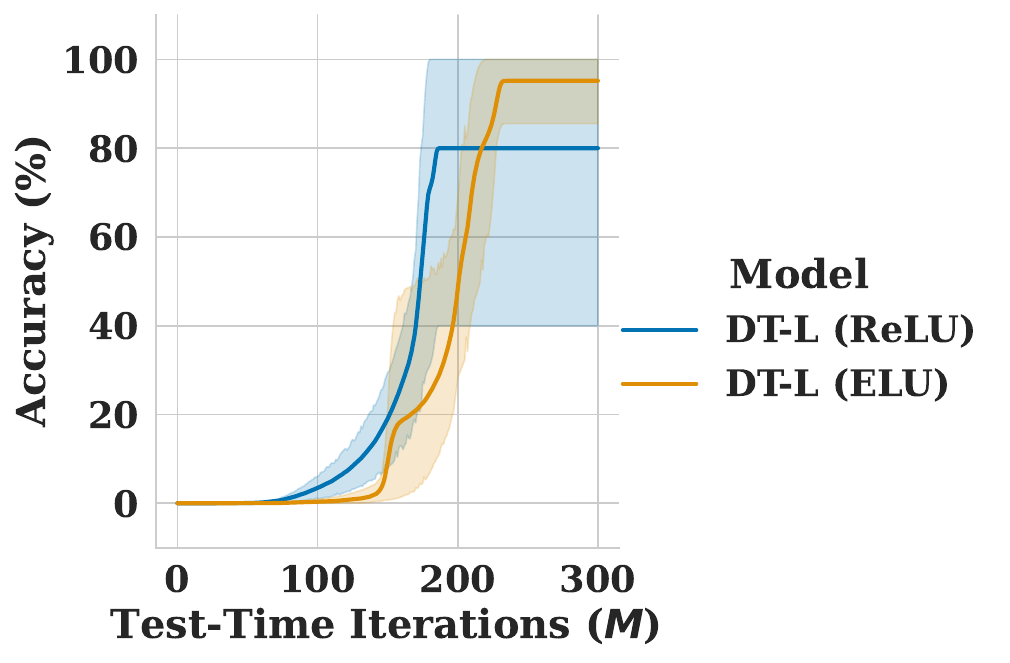}\\
        (a)&(b)
    \end{tabular}
    \captionof{figure}{Mean solution accuracy at different \(M\) for (a) 5
                       instances of \DTR{} (with ReLU activations) and 5
                       instances of \DTR{} (with ELU activations), and for (b)
                       5 instances of \DTL{} (with ReLU activations) and 5
                       instances of \DTL{} (with ELU activations). All models
                       have width \(w=32\). Extrapolation performance tested on
                       512-bit prefix sum problem instances. Shaded areas show
                       95\% confidence intervals.}
    \label{fig:ablation:activation}
\end{figure}

\paragraph{Activation Function.}
We perform a study by modifying \DTR{} to use ELU \citep{clevert2016elu}
instead of ReLU. From \cref{fig:ablation:activation}(a) it appears that ELU
allows for increased extrapolation performance from \DTR{} with minimal decay.

Similarly, we perform a study by modifying \DTL{} to use ReLU instead of
ELU. It can be seen from \cref{fig:ablation:activation}(b) that, while ReLU
results in faster convergence, ELU appears to provide overall increased
extrapolation performance.

\paragraph{Bias in the Final Layer.}
\DTR{} has no bias terms throughout the entire model. We compare this to a
version which has a bias term in \textit{only} the final layer of the model.

\begin{figure}[H]
    \centering
    \begin{tabular}{cc}
        \includegraphics[scale=0.33]{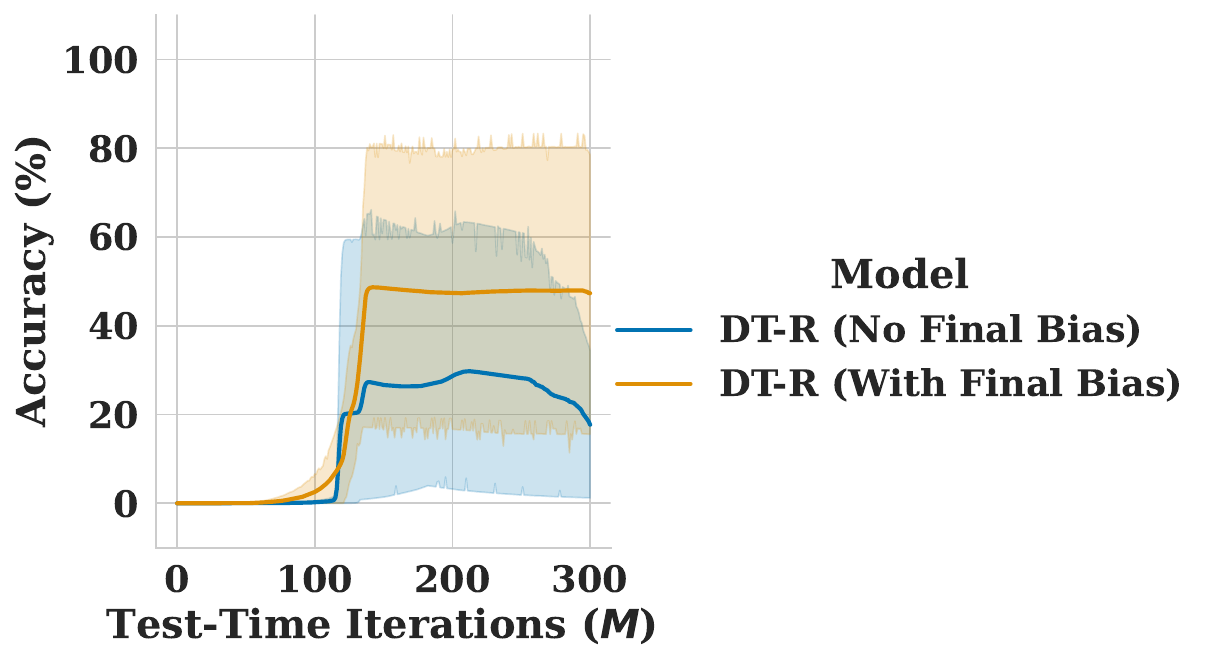}&
        \includegraphics[scale=0.33]{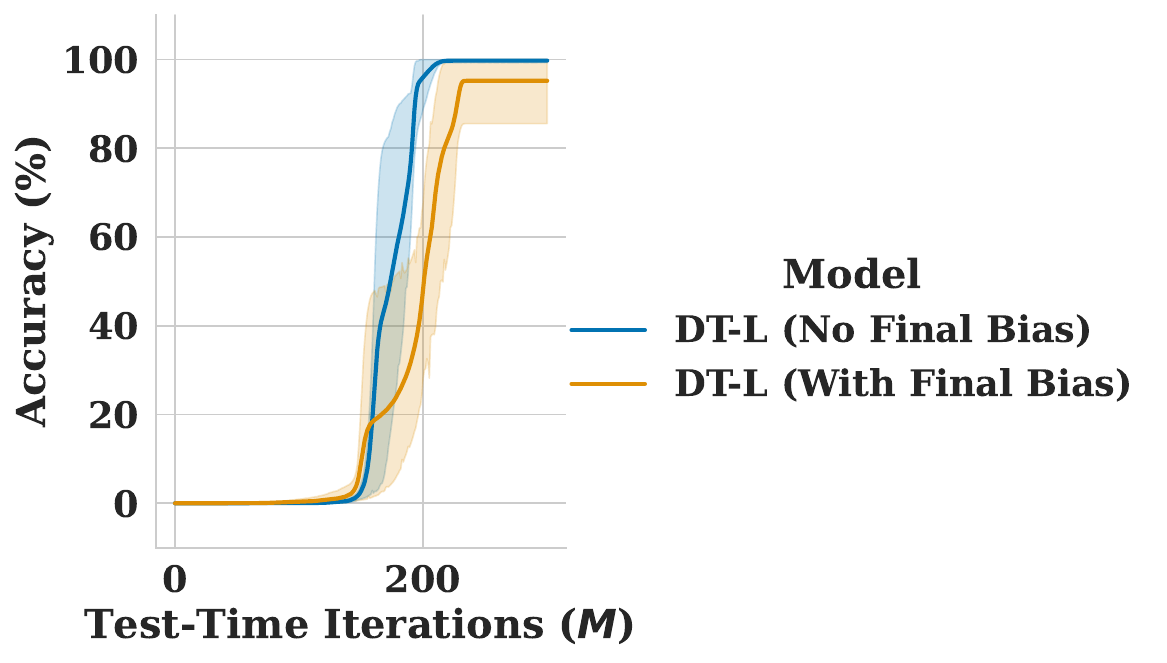}\\
        (a)&(b)
    \end{tabular}
    \captionof{figure}{Mean solution accuracy at different \(M\) for (a) 5
                       instances of \DTR{} (without final bias term) and 5
                       instances of \DTR{} (with final bias term), and for (b)
                       5 instances of \DTL{} (without final bias term) and 5
                       instances of \DTL{} (with final bias term). All models
                       have width \(w=32\). Extrapolation performance tested on
                       512-bit prefix sum problem instances. Shaded areas show
                       95\% confidence intervals.}
    \label{fig:ablation:finalbias}
\end{figure}

\begin{figure}[H]
    \centering
    \begin{tabular}{cc}
        \includegraphics[scale=0.33]{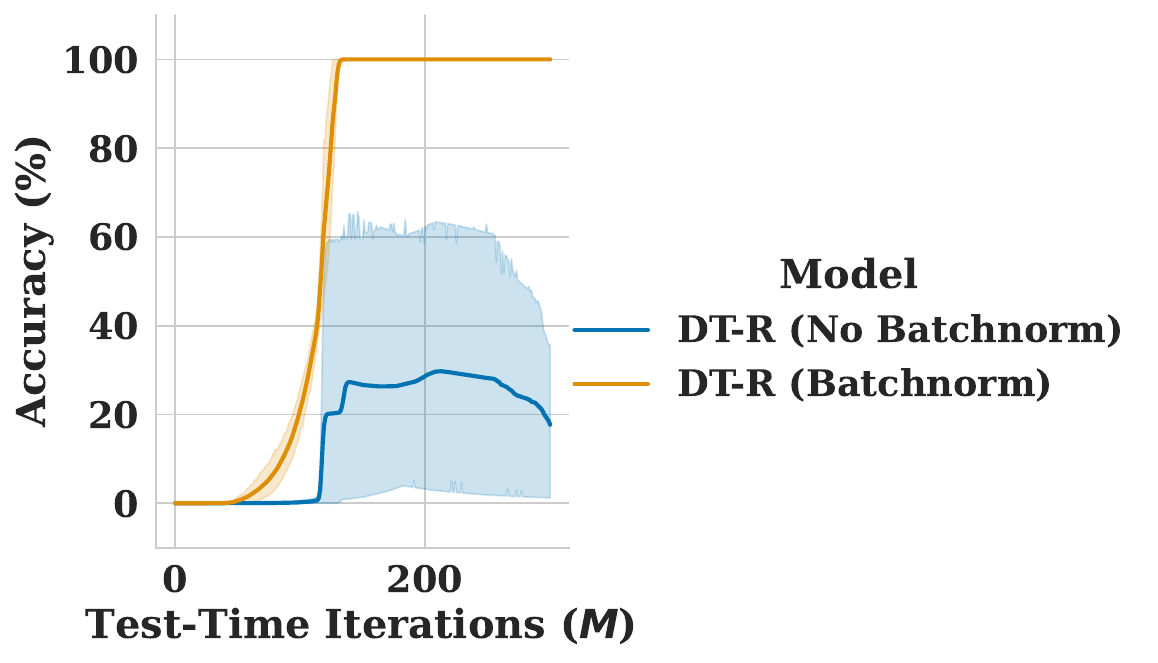}&
        \includegraphics[scale=0.33]{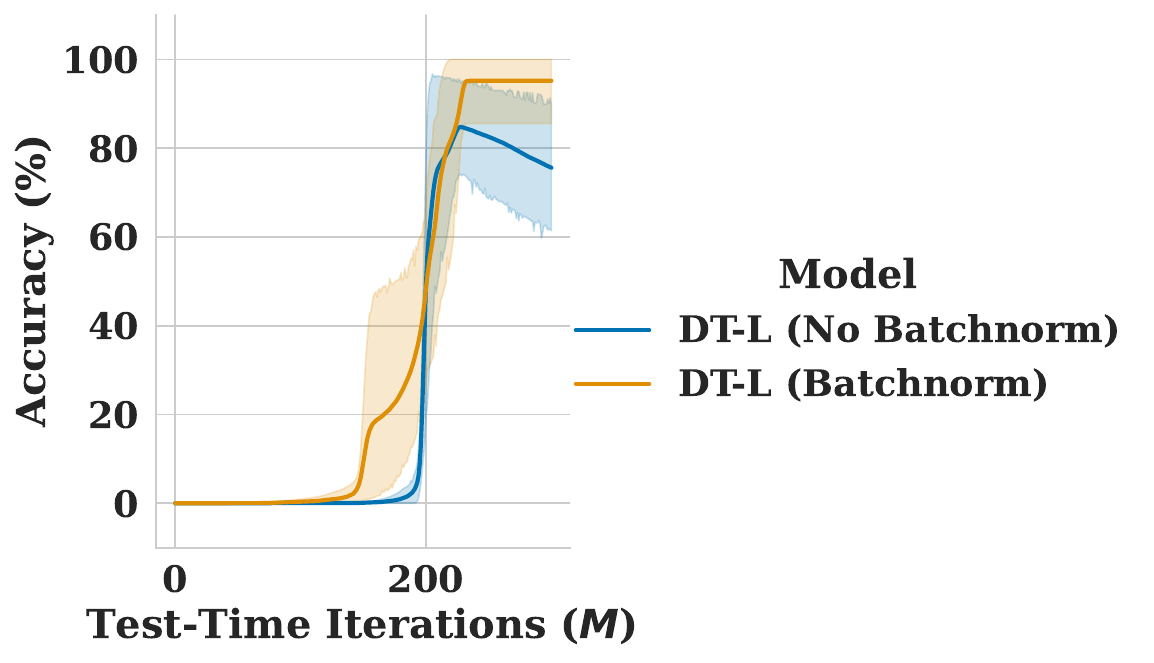}\\
        (a)&(b)
    \end{tabular}
    \captionof{figure}{Mean solution accuracy at different \(M\) for (a) 5
                       instances of \DTR{} (without batch normalization) and 5
                       instances of \DTR{} (with batch normalization), and for (b)
                       5 instances of \DTL{} (without batch normalization) and 5
                       instances of \DTL{} (with batch normalization). All models
                       have width \(w=32\). Extrapolation performance tested on
                       512-bit prefix sum problem instances. Shaded areas show
                       95\% confidence intervals.}
    \label{fig:ablation:batchnorm}
\end{figure}

\paragraph{Batch Normalization.}
\DTR{} has no batch normalization layers in the model. We perform a study to
measure the impact of adding batch normalization to \DTR{} (\cref{fig:ablation:batchnorm}(a)), as well as
remove it from \DTL{} (\cref{fig:ablation:batchnorm}(b)).

\section{Running Times and Peak Memory Usage}
\label{app:runtimes}

After all experiments had been completed, a modified version of the
spectral normalization code was created which caches the normalized weights.
This modification gives improved training times.

\cref{tab:my_table} provides the training times of \DTR{} and \DTL{}
\textit{before} this modification (the version used in the experiments).
\cref{tab:runtimenew} provides the training times of \DTL{} \textit{after}
this modification.

\begin{table}[H]
    \centering
    \caption{Representative training times and peak memory usage for different models. Training time is given as \texttt{hours:minutes}.}
    \begin{tabular}{lllllll}
    \toprule
         &  & Batch &   &   & Training & Memory \\ 
        Model & Problem & Size & Epochs & Hardware & Time & Usage \\ \midrule
        \DTR{} \(w=32\) & Prefix Sums & 500 & 150 & RTX8000 & 00:10 & 1.25 GB \\
        \DTL{} \(w=32\) & Prefix Sums & 500 & 150 & RTX8000 & 00:30 & 1.34 GB \\ 
        \DTR{} \(w=32\) & Mazes & 50 & 50 & RTX8000 & 03:30 & 4.30 GB \\
        \DTL{} \(w=32\) & Mazes & 50 & 50 & RTX8000 & 04:02 & 4.41 GB \\
        \DTR{} \(w=16\) & Chess Puzzles & 300 & 120 & A100 & 02:55 & 10.87 GB \\
        \DTL{} \(w=16\) & Chess Puzzles & 300 & 120 & RTX8000 & 07:36 & 10.67 GB \\
        \bottomrule
    \end{tabular}
    \label{tab:my_table}
\end{table}

\begin{table}[H]
    \centering
    \caption{Representative training times and peak memory usage for different models (using spectral normalized weight caching). Training time is given as \texttt{hours:minutes}.}
    \begin{tabular}{lllllll}
    \toprule
         &  & Batch &   &   & Training & Memory \\ 
        Model & Problem & Size & Epochs & Hardware & Time & Usage \\ \midrule
        \DTL{} \(w=32\) & Prefix Sums & 500 & 150 & RTX8000 & 00:19 & 1.30 GB \\ 
        \DTL{} \(w=32\) & Mazes & 50 & 50 & RTX8000 & 03:15 & 4.42 GB \\
        \DTL{} \(w=16\) & Chess Puzzles & 300 & 120 & RTX8000 & 04:47 & 10.70 GB \\
        \bottomrule
    \end{tabular}
    \label{tab:runtimenew}
\end{table}

\clearpage 
\section*{NeurIPS Paper Checklist}

\begin{enumerate}

\item {\bf Claims}
    \item[] Question: Do the main claims made in the abstract and introduction accurately reflect the paper's contributions and scope?
    \item[] Answer: \answerYes{} 
    \item[] Justification: We claim an improvement to an existing type of model known as a \dt{} network as a result of careful analysis of failure cases. Empirically we show that our improved \dtl{} (\DTL{}) model has better training and extrapolation performance. We also show a theoretical proof of why our proposed changes guarantee convergence of the learned algorithm (and why the earlier approaches did not). 

\item {\bf Limitations}
    \item[] Question: Does the paper discuss the limitations of the work performed by the authors?
    \item[] Answer: \answerYes{} 
    \item[] Justification: We have discussed limitations in some detail and created a separate limitations paragraph in \cref{sec:discussion}. The approach we develop analytically has relatively few direct limitations as the assumptions are mild. However, even on simple problems we recognise that whilst we have made significant improvements over prior approaches there are still times where the model fails to learn for example. 

\item {\bf Theory Assumptions and Proofs}
    \item[] Question: For each theoretical result, does the paper provide the full set of assumptions and a complete (and correct) proof?
    \item[] Answer: \answerYes{} 
    \item[] Justification: Sketch-proofs and the intuition behind our architectural designs of a recurrent function that converges are given in the body of the paper, with more formal proofs in the appendices (\cref{app:proofs}) and are cross-referenced. All assumptions are clearly stated.

    \item {\bf Experimental Result Reproducibility}
    \item[] Question: Does the paper fully disclose all the information needed to reproduce the main experimental results of the paper to the extent that it affects the main claims and/or conclusions of the paper (regardless of whether the code and data are provided or not)?
    \item[] Answer: \answerYes{} 
    \item[] Justification: Full details of our model design is given in \cref{app:architecture}, with training details provided in \cref{app:training}. The main evaluation setup repeats that described by \citet{bansal2022endtoend} and \citet{schwarzschild2021algorithm} where we use the datasets provided by \citet{schwarzschild2021dataset} for studying easy-to-hard generalisation. Further we provide the code for all models and experiments (as supplementary material during review, and later as a GitHub link).

\item {\bf Open access to data and code}
    \item[] Question: Does the paper provide open access to the data and code, with sufficient instructions to faithfully reproduce the main experimental results, as described in supplemental material?
    \item[] Answer: \answerYes{} 
    \item[] Justification: All code for reproducing experiments is provided (in the supplementary material during review; to be moved to github later), including code for generating our TSP datasets. As described above we use the open access datasets provided by \citet{schwarzschild2021dataset} for studying easy-to-hard generalization for the main evaluation.

\item {\bf Experimental Setting/Details}
    \item[] Question: Does the paper specify all the training and test details (e.g., data splits, hyperparameters, how they were chosen, type of optimizer, etc.) necessary to understand the results?
    \item[] Answer: \answerYes{} 
    \item[] Justification: Please see \cref{app:training} for full details of the training procedure and \cref{app:architecture} for problem-specific model architectures. The broader experimental setting can be found in \cref{sec:results} and \cref{sec:tsp_results} of the main paper.
\newpage
\item {\bf Experiment Statistical Significance}
    \item[] Question: Does the paper report error bars suitably and correctly defined or other appropriate information about the statistical significance of the experiments?
    \item[] Answer: \answerYes{} 
    \item[] Justification: Error bars (confidence intervals) are provided where appropriate. Because part of the objective of the paper is to demonstrate the improvement in stability of the models, in some cases we show data from multiple raw runs of the models, thus allowing different cases to be seen.

\item {\bf Experiments Compute Resources}
    \item[] Question: For each experiment, does the paper provide sufficient information on the computer resources (type of compute workers, memory, time of execution) needed to reproduce the experiments?
    \item[] Answer: \answerYes{} 
    \item[] Justification: All experiments are carried out on commodity hardware, where we used GPU resources that we had available. This is documented in \cref{sec:results} and \cref{sec:tsp_results}, with a fuller table of exemplar configurations and timings for different experiments in \cref{app:runtimes}.
    
\item {\bf Code Of Ethics}
    \item[] Question: Does the research conducted in the paper conform, in every respect, with the NeurIPS Code of Ethics \url{https://neurips.cc/public/EthicsGuidelines}?
    \item[] Answer: \answerYes{} 
    \item[] Justification: We have studied the code of ethics and can confirm our research conforms to all guidelines. Our research uses no human participants, and the datasets we experiment with are created artificially to control biases. Our research does not pose any immediate security or safety issues, however in body of the paper we recognise that their are potential broader impacts from the approaches we are building. The approaches we develop in the paper are aimed at being more stable and reliable, whilst also potentially using fewer parameters. As such they are promoting positive environmental impact. 

\item {\bf Broader Impacts}
    \item[] Question: Does the paper discuss both potential positive societal impacts and negative societal impacts of the work performed?
    \item[] Answer: \answerYes{} 
    \item[] Justification: We have included a discussion of potential broader impacts in \cref{sec:discussion}. Summarising, whilst this work is at the fundamental principles end of the spectrum, we see many potential applications from a learning machine that can learn iterative algorithms that extrapolate. Whilst the work does not directly have negative societal impact, it is quite conceivable that an application that does could be built using it as the basis.
    
\item {\bf Safeguards}
    \item[] Question: Does the paper describe safeguards that have been put in place for responsible release of data or models that have a high risk for misuse (e.g., pretrained language models, image generators, or scraped datasets)?
    \item[] Answer: \answerNA{} 
    \item[] Justification: The model architectures and datasets as presented offer minimal potential for misuse.

\item {\bf Licenses for existing assets}
    \item[] Question: Are the creators or original owners of assets (e.g., code, data, models), used in the paper, properly credited and are the license and terms of use explicitly mentioned and properly respected?
    \item[] Answer: \answerYes{} 
    \item[] Justification: Where we have used existing code or implementations these are properly attributed. The datasets~\citep{schwarzschild2021dataset} we use are referenced in the paper. The papers that develop model architectures~\citep{schwarzschild2021algorithm, bansal2022endtoend} we build upon are referenced in the paper.

\item {\bf New Assets}
    \item[] Question: Are new assets introduced in the paper well documented and is the documentation provided alongside the assets?
    \item[] Answer: \answerYes{} 
    \item[] Justification: Our code and TSP dataset generator will be released on github, and are provided in the supplementary material during review.

\item {\bf Crowdsourcing and Research with Human Subjects}
    \item[] Question: For crowdsourcing experiments and research with human subjects, does the paper include the full text of instructions given to participants and screenshots, if applicable, as well as details about compensation (if any)? 
    \item[] Answer: \answerNA{} 
    \item[] Justification: This work does not involve crowdsourcing or human participants.

\item {\bf Institutional Review Board (IRB) Approvals or Equivalent for Research with Human Subjects}
    \item[] Question: Does the paper describe potential risks incurred by study participants, whether such risks were disclosed to the subjects, and whether Institutional Review Board (IRB) approvals (or an equivalent approval/review based on the requirements of your country or institution) were obtained?
    \item[] Answer: \answerNA{} 
    \item[] Justification: This work does not involve crowdsourcing or human participants.

\end{enumerate}

\end{document}